\documentclass[lettersize,journal]{IEEEtran}
\usepackage{amsmath,amsfonts}
\usepackage{algorithmic}
\usepackage{algorithm}
\usepackage{array}
\usepackage[caption=false,font=normalsize,labelfont=sf,textfont=sf]{subfig}
\usepackage{textcomp}
\usepackage{stfloats}
\usepackage{url}
\usepackage{verbatim}
\usepackage{graphicx}
\usepackage{cite}
\usepackage{amssymb}
\usepackage{bm}
\usepackage{bbm}
\usepackage{pifont}
\usepackage{adjustbox}
\usepackage{booktabs}
\usepackage{multirow}
\usepackage{multicol}
\usepackage{rotating}
\begin{document}

\title{Auto-FlexSwitch: Efficient Dynamic Model Merging via Learnable Task Vector Compression}

\author{Junqi Gao\textsuperscript{1}, Dazhi Zhang\textsuperscript{1}, Zhichang Guo\textsuperscript{1,\dag}, Biqing Qi\textsuperscript{2}, Yi Ran\textsuperscript{1}, Wangmeng Zuo\textsuperscript{1}
        
\thanks{
\textsuperscript{1}School of Mathematics, Harbin Institute of Technology, Harbin, P. R. China;
\textsuperscript{2}Shanghai Artificial Intelligence Laboratory, Shanghai, P. R. China.
\textsuperscript{\dag} Corresponding author: Zhichang Guo.\\
Emails: gjunqi97@gmail.com; zhangdazhi@hit.edu.cn; mathgzc@hit.edu.cn; qibiqing7@gmail.com; yi.ran@hit.edu.cn; cswmzuo@gmail.com.

}

}

\maketitle

\begin{abstract}
Model merging has attracted attention as an effective path toward multi-task adaptation by integrating knowledge from multiple task-specific models. Among existing approaches, dynamic merging mitigates performance degradation caused by conflicting parameter updates across tasks by flexibly combining task-specific parameters at inference time, thereby maintaining high performance. However, these methods require storing independent parameters for each task, resulting in prohibitive storage overhead. To address this issue, we first experimentally demonstrate that the fine-tuned weight increments (referred to as task vectors) exhibit an impulse-like activation pattern and high robustness to low-bit representations. Driven by this insight, we propose T-Switch, which decomposes task vectors into three compact components: a binary sparse mask (indicating activation positions), a sign vector (representing parameter polarity), and a scalar scaling factor, achieving high-fidelity approximation at high compression ratios. We then introduce Auto-Switch, a training-free merging scheme that automatically composes task vectors via feature similarity retrieval. Building on this, we develop Auto-Switch, a training-free merging scheme that automatically assembles task vectors through feature similarity retrieval. Furthermore, to transform task vector sparsification and quantization from static rules to adaptive learning, we propose FlexSwitch, a learnable framework which jointly optimizes the compression strategy for each model unit via \textbf{L}earnable \textbf{G}ating \textbf{S}parsification (LGS) and \textbf{B}it-width \textbf{A}daptive \textbf{S}election (BAS), while employing the \textbf{S}parsity-\textbf{A}ware \textbf{S}torage \textbf{S}trategy (SASS) to select the optimal storage encoding structure. Finally, by incorporating a \textbf{K}-\textbf{N}earest \textbf{N}eighbor (KNN) inference scheme with a learnable low-rank metric, we present Auto-FlexSwitch, a dynamic model merging approach that supports highly efficient task vector compression. Experiments across diverse model architectures and downstream benchmarks demonstrate its strong performance with substantial storage savings.
\end{abstract}


\section{Introduction}
\IEEEPARstart{W}{ith} the rapid growth of various open-source communities \cite{abs-1910-03771,abs-1906-07155}, an increasing number of pre-trained and fine-tuned models are widely shared \cite{abs-2505-09388,abs-2503-19786,abs-2412-08905}. Deploying these models directly to address specific tasks has become a mainstream practice. Yet, in the face of diverse task scenarios, maintaining a dedicated model for each task incurs prohibitive storage and deployment overhead, rendering it impractical, especially in resource-constrained contexts. To address this challenge, model merging \cite{IlharcoRWSHF23,MatenaR22} offers a promising solution. By integrating the parameters of multiple task-specific models into a single unified model, knowledge from different tasks can be effectively consolidated, thereby enhancing the model’s adaptability and performance in multi-task settings.

Model merging approaches can be broadly categorized into two types. The first is static merging \cite{MatenaR22,Jin0P023,IlharcoRWSHF23}, where the model weights remain fixed during inference. However, due to distributional discrepancies among different tasks, the directions of the task-specific incremental weights (also known as task vectors) often conflict with one another, limiting the merged model’s performance across individual tasks. Consequently, a single static merged model struggles to adapt to dynamically changing task data. In contrast, dynamic merging methods \cite{lu2024twin,huang2024emr} flexibly adjust the merging strategy by dynamically composing task vectors based on the input, enabling the model to adaptively emphasize the relevant task vector in response to varying tasks, thereby naturally alleviating directional conflicts among task vectors. Nevertheless, these methods introduce a new trade-off: they typically require maintaining task-specific parameter components (such as task vectors or masks), leading to cumulative storage overhead. Therefore, achieving a better trade-off between performance and storage efficiency has become critical to the broader practical adoption of dynamic merging approaches. Against this backdrop, our work aims to develop a dynamic merging framework that simultaneously achieves high performance and high storage efficiency, offering an effective solution to the aforementioned challenges.

To achieve this goal, a natural approach is to efficiently compress the task vectors maintained for each task without compromising performance. To this end, we explore high-efficiency compression and storage mechanisms for task vectors through the lenses of sparsification and quantization, upon which we build a high-performance dynamic merging method. However, the efficacy of this compression scheme hinges on a critical premise: task vectors possess an inherent structural capacity for sparsification and quantization. We begin by conducting an experimental analysis to verify this premise.

Regarding sparsifiability, we systematically investigate the relationship between the magnitude of parameters in the task vector and their contribution to the target task. Specifically, we find that the parameters exhibit an impulse-like activation pattern: only those whose magnitudes exceed a certain activation threshold make a significant contribution to the task. Interestingly, pruning the remaining parameters not only preserves task accuracy but can even lead to further performance gains. As the pruning ratio increases, this performance improvement shows a continuous upward trend, only beginning to decay after the ratio exceeds $70\%$. This phenomenon provides strong evidence that task vectors possess an inherent structural capacity for sparsification.

We further explore the quantizability of the sparse task vectors obtained after magnitude-based pruning. Specifically, by replacing non-zero parameters with their signs (binary quantization) and restoring the original vector scale via L2-norm scaling, we observe that the model's performance on target tasks is effectively maintained. Notably, as the sparsity increases, the performance gap between the binarized and the original sparse task vectors consistently narrows. This suggests that highly sparse task vectors are robust to magnitude precision, where binary sign information alone suffices to sustain performance, thus validating their intrinsic quantizability.

Building on these findings, we propose Task Switch (T-Switch), a method for constructing lightweight task vectors. It explicitly decomposes the task vector into three compact components: a switch knob formed by a single scaling factor, an activation switch instantiated by a binarized mask vector that specifies the sparsity rate, and a polarity switch instantiated by a binarized sign vector. This decomposition enables an extremely compact and efficient encoding of task-specific knowledge while preserving the original performance, theoretically yielding at least a 16$\times$ reduction in storage overhead. Based on this, we further introduce Auto-Switch, a training-free dynamic merging approach. It constructs a query set using the features from a small number of instances of the target task, and performs similarity retrieval over this set based on the input features to compute the combination weights for T-Switch, thereby enabling task-adaptive model merging.

Inspired by the three-component structure of T-Switch, we further contemplate: \emph{Can the construction for lightweight task vectors be transformed from \textbf{relying on fixed rules} (such as a preset sparsity rate, hard binarization, and static scaling) into an \textbf{adaptive construction mechanism driven by end-to-end optimization}, thereby automatically learning the optimal sparsity level, quantization bit-width, and magnitude calibration for the task vectors of each model module?}

Driven by this vision, we first design \textbf{L}earnable \textbf{G}ating \textbf{S}parsification (LGS). This method equips task vectors of each model module with a learnable magnitude threshold and employs a temperature-controlled sigmoid function to generate continuous gating signals, enabling differentiable sparsification, while a learnable scaling factor is introduced to calibrate the overall magnitude after sparsifying. Validation experiments demonstrate that LGS can achieve performance comparable to, or even surpassing, full-parameter fine-tuning at an average sparsity rate of $97\%$. Subsequently, we introduce \textbf{B}it-width \textbf{A}daptive \textbf{S}election (BAS) to allocate the optimal quantization bit-width for each model module's task vector. Furthermore, to maximize storage efficiency, we propose a \textbf{S}parsity-\textbf{A}ware \textbf{S}torage \textbf{S}trategy (SASS). It employs a grouped COO format for the task vector parameters and adaptively selects the optimal group configuration based on the actual sparsity ratio, thereby maximizing the utilization of the compressed task vector's sparsity to achieve smaller storage footprint.

Building on these component designs, we organically integrate LGS, BAS, and SASS to propose the FlexSwitch framework for constructing learnable lightweight task vectors. FlexSwitch jointly optimizes the sparse structure and quantization bit-width of each model module via LGS and BAS while maintaining performance, and automatically selects the optimal storage encoding structure using SASS to minimize actual storage overhead. In contrast to T-Switch’s reliance on fixed sparsity, hard binarization, and static scaling, FlexSwitch achieves higher compression and better task performance. Finally, we introduce a \textbf{K}-\textbf{N}earest \textbf{N}eighbor (KNN) inference mechanism with a learnable low-rank metric to enable more accurate retrieval for lightweight task vectors, forming Auto-FlexSwitch, an adaptive merging scheme that supports highly efficient task vector compression.

To comprehensively evaluate the proposed method, we verify its advantages in balancing merging performance with low storage overhead on multi‑task benchmarks across various model architectures and downstream scenarios. Additionally, we assess the potential of FlexSwitch as a lightweight storage strategy for fine‑tuned Large Language Model (LLM) weights.

Compared to our previous conference version \cite{QiLWGL0025}, this paper extends and improves the work in the following aspects: (i) proposes FlexSwitch, a lightweight task vector framework that adaptively sparsifies and quantizes task vectors of different modules via LGS and BAS, jointly optimizing sparsity and bit-width while preserving performance; (ii) designs SASS, a storage strategy for sparsified-quantized task vectors that stores each module's task vector in a grouped COO format and adaptively selects the optimal group count based on the actual sparsity ratio, maximally exploiting sparsity to reduce storage; (iii) further proposes Auto-FlexSwitch, a dynamic merging scheme that integrates a K-nearest neighbor inference mechanism with a learnable low-rank metric, achieving accurate task vector assignment with lower retrieval overhead; (iv) conducts more systematic evaluations across broader model architectures and more diverse downstream scenarios to more comprehensively verify the effectiveness of the proposed methods.

The main contributions of this work are as follows:
\begin{itemize}
    \item Through controlled experiments, we reveal the impulse-like activation pattern of task vectors and their high tolerance for low-bit representations, offering critical insights for designing efficient task vector compression methods.
    \item Building on these observations, we design T‑Switch, a simple method that constructs lightweight task vectors using three compact components. Furthermore, we propose Auto-Switch, a dynamic merging scheme that automates merging via retrieval from a small query set.
    \item We propose FlexSwitch, a framework that jointly optimizes the sparsity structure and quantization bit-width of task vectors through learnable gating and adaptive bit-width selection. To our knowledge, this is the first learnable framework simultaneously optimizing sparsity and bit-width allocation for task vectors.
    \item We design SASS, which automatically selects the optimal group configuration from different group count settings of the grouped COO format based on the actual sparsity ratio of each module's task vector, thereby fully exploiting the sparsity for storage efficiency.
    \item We propose Auto-FlexSwitch, a dynamic merging method integrating a KNN mechanism with a learnable low-rank metric for more accurate and efficient task vector assignment. Its effectiveness is extensively validated across diverse model architectures and downstream scenarios.
\end{itemize}

\section{Related Works}
\subsection{Model Merging}
Model merging aims to integrate multiple task-specific fine-tuned models into a unified model, providing an efficient solution for multi-task adaptation. Based on whether the merging strategy adjusts dynamically according to input samples, existing research can be broadly categorized into static merging and dynamic merging.

Static merging methods assume that the merged weights remain invariant during inference. Early research, grounded in Linear Mode Connectivity theory \cite{DraxlerVSH18,GaripovIPVW18,FrankleD0C20}, demonstrated the feasibility of consolidating knowledge from multiple models via weight interpolation. This catalyzed a series of subsequent developments, ranging from early attempts like weight averaging \cite{WortsmanIGRLMNF22} and task arithmetic \cite{IlharcoRWSHF23} to more sophisticated reweighting strategies. Notable examples include Fisher-information-based weighted averaging \cite{MatenaR22,LeeLWWCW25}, formulating merging as an optimization problem \cite{AinsworthHS23,Jin0P023,YangW00G0T24}, and feature matching guided merging \cite{WangYSPK20,StoicaBBRHH24}. However, static merging is inherently susceptible to interference due to conflicting parameter update directions across different tasks. To mitigate this, methods such as DARE \cite{Yu0Y0L24} and Ties-Merging \cite{YadavTCRB23} explore sparsification to alleviate inter-task conflicts; nevertheless, static merging still struggles to fully bypass the interference caused by directional misalignments. 

In contrast, dynamic merging effectively mitigates inter-task conflicts by adaptively combining task-specific parameters based on input samples. Such approaches either implement flexible knowledge reorganization via task masks \cite{huang2024emr} or task-specific components \cite{lu2024twin} on carefully constructed shared weights, or partition layers for static and dynamic merging based on feature similarity \cite{YeHSCHO25}. While these methods enhance multi-task performance, they necessitate maintaining expert parameters for each task, leading to substantial storage overhead. Therefore, how to preserve the advantages of dynamic merging while drastically reducing storage overhead through efficient compression and ensuring stability across varying intensities remains a critical open problem.

\subsection{Fine-tuned Weight Compression}
As the scale of model parameters continues to grow, the storage overhead of fine-tuned weights has become increasingly prominent. Sparsification and quantization, as two mainstream techniques for weight compression, have been widely explored. Early research promoted incremental weight sparsification during the fine-tuning process by directly incorporating sparsification regularization \cite{GuoRK20,AnsellPKV22,HuZDWWLS22}. Other works have introduced indicative metrics to guide sparsification, such as leveraging Fisher information \cite{SungNR21}, gradient information \cite{BhardwajPPGK0BW24}, or feature activations \cite{HeLJM25} to measure the importance of parameters across different positions. The above approaches primarily perform compression during the training process. In contrast, some studies have attempted to directly apply quantization approximations to post-training fine-tuned weights \cite{NingW0Z0CLZ24,LiuXLL0DC24}. However, the quantization methods employed in these studies are often direct and fixed, lacking adaptive bit-rate allocation for different modules based on the specific characteristics of the fine-tuned weights. More importantly, they do not explore the synergistic advantages of combining quantization with sparsification, nor has a joint sparsification-quantization compression technique been established within the context of model merging.

\section{Proposed Method}
\subsection{Problem Formulation}
Given a pre-trained model $f(\cdot ; \bm{\Theta})$ parameterized by weights $\bm{\Theta}=\left\{\bm{\theta}^{l}\right\}_{l=1}^{L}$, where $\bm{\theta}^{l} \in \mathbb{R}^{n_l}$ denotes the weight vector of the $l$-th module and $\sum_{l=1}^L n_l = n$ represents the total number of the model parameters. Consider a set of tasks $\{\mathcal{T}_{k}\}_{k=1}^{K}$. For each task $\mathcal{T}_k$, the corresponding fine-tuned model parameters are denoted as $\bm{\Theta}_k = \{\bm{\theta}_k^l\}_{l=1}^L$. The task vector for task $\mathcal{T}_k$ is then computed as $\bm{\tau}_k = \{ \bm{\tau}_k^l \mid \bm{\tau}_k^l = \bm{\theta}_k^l - \bm{\theta}^l \}_{l=1}^L$. The objective of model merging is to combine the pre-trained model with the set of task vectors $\{\boldsymbol{\tau}_{k}\}_{k=1}^{K}$ to obtain a merged model that performs well across all tasks, formulated as:
\begin{equation} 
\min_{\mathcal{M}} \mathbb{E}_{(\bm{x},y)\in\cup_{k=1}^K\mathcal{T}_{k}} \ell \left( f \left( \bm{x}; \mathcal{M} \left( \bm{\Theta}, \{\bm{\tau}_{k}\}_{k=1}^K \right) \right), y \right), 
\end{equation}
where $(\bm{x}, y) \in \cup_{k=1}^K\mathcal{T}_{k}$ denotes an input-label pair from the joint distribution of all tasks, and $\mathcal{M}$ is the merging operation, ranging from simple linear combinations \cite{IlharcoRWSHF23} to more complex merging after certain preprocessing of the task vectors \cite{Yu0Y0L24}.

\begin{figure*}[t]
    \centering
    \includegraphics[width=0.88\textwidth]{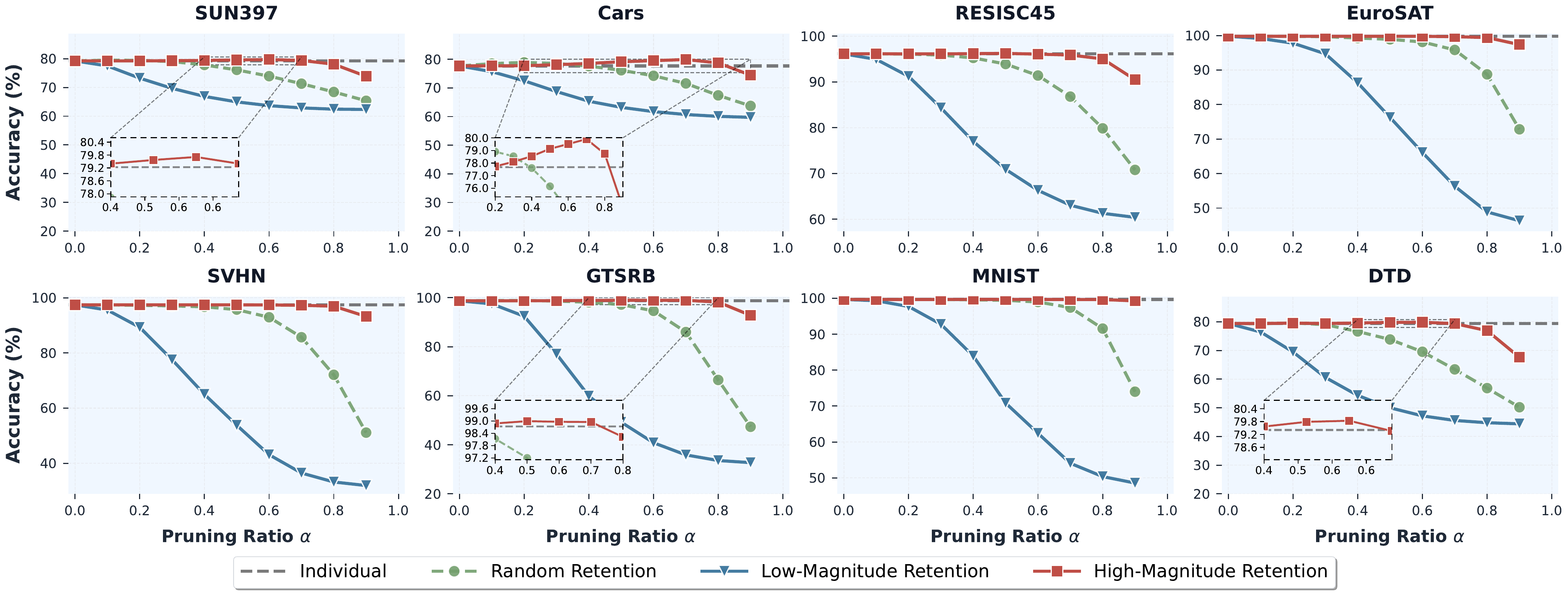}
    \vspace{-7pt}
    \caption{Accuracy ($\%$) trends of the three control strategies \textbf{C1}, \textbf{C2}, and \textbf{C3} across the eight visual tasks on the ViT-B/32 model as a function of the pruning rate $\alpha$. The horizontal dashed lines (Individual) represent the original fine-tuning accuracy for each task. The insets highlight the regions where specific tasks exhibit performance substantially exceeding the fine-tuning baseline (by more than $0.2\%$).}
    \label{fig:valid_exp}
    \vspace{-12pt}
\end{figure*}

\subsection{Efficient Dynamic Merging with Lightweight Task Switches}

\subsubsection{The Sparsifiability of Task Vectors}
\label{sec:p-spar}
We first explore the sparsity properties of task vectors by examining the relationship between parameter magnitude and its contribution to the target task. Intuitively, parameters that undergo significant shifts during fine-tuning are more likely to carry critical task-specific information. Moreover, prior research has established the efficacy of magnitude-based metrics for assessing parameter importance \cite{ZhuG18,Sun0BK24}. To validate the applicability of this criterion to task vectors, we design the following impulse activation function to retain parameters within different magnitude ranges in a task vector:
\begin{equation}
    g_{m}({\tau}^l_{k,j})=\left\{\begin{matrix}
 1, & \text{if } {\tau}^l_{k,j}>\gamma^l_{+} \text{ or } {\tau}^l_{k,j}<\gamma_{-}^l \\
  0 & \text{else} 
\end{matrix}\right.,
\end{equation}
where $\gamma_{+}^l>0$ and $\gamma_-^l<0$ denote the upper and lower activation thresholds for the task vector components of the $l$-th module, respectively, and ${\tau}^l_{k,j}$ represents the $j$-th element of the task vector $\bm{\tau}^l_k$. Building upon this, we define the global impulse activation operator as $\mathcal{G}_m(\bm{\tau}_k) \triangleq \{ \bm{\tau}_k^l \odot \mathbf{g}_{m}(\bm{\tau}^l_{k}) \}_{l=1}^L$ and $\mathbf{g}_{m}(\bm{\tau}^l_{k})\in \{0, 1\}^{n_l}$ is a module-level mask whose elements are given by $(\mathbf{g}_{m}(\bm{\tau}^l_{k}))_j = g_m({\tau}_{k,j}^l)$. We then conduct experiments using CLIP-ViT-B/32 \cite{RadfordKHRGASAM21} as the backbone across a multi-task benchmark comprising eight visual tasks: SUN397 \cite{XiaoHEOT10}, Cars \cite{Krause0DF13}, RESISC45 \cite{ChengHL17}, EuroSAT \cite{HelberBDB19}, SVHN \cite{netzer2011reading}, GTSRB \cite{StallkampSSI11}, MNIST \cite{Deng12}, and DTD \cite{CimpoiMKMV14}. By varying the pruning rate $\alpha$, we investigate the impact of task vector parameters within different magnitude ranges on performance. Specifically, we establish three control groups as follows:

\noindent\textbf{C1 (High-Magnitude Retention)}: Let $\gamma_+^l$ and $\gamma_-^l$ be the $\alpha$-quantiles of the positive and negative elements of $\bm{\tau}_k^l$, respectively, thereby pruning the $\alpha\times 100\%$ proportion of elements with the smallest magnitudes and retaining only the task vector parameters with larger absolute values. The activated task vector is denoted as $\hat{\bm{\tau}}_k = \mathcal{G}_m(\bm{\tau}_k)$. 

\noindent\textbf{C2 (Low-Magnitude Retention)}: In contrast to C1, this group employs the mask $(\mathbf{1} - \mathbf{g}_{m}(\bm{\tau}^l_{k}))$ to prune high-magnitude parameters, preserving only the low-magnitude elements within the interval $[\gamma_-^l, \gamma_+^l]$, i.e., $\hat{\bm{\tau}}_k = \{ \bm{\tau}_k^l \odot (\mathbf{1} - \mathbf{g}_{m}(\bm{\tau}^l_{k})) \}_{l=1}^L$. 

\noindent\textbf{C3 (Random Retention)}: Elements of the task vector are randomly pruned at a rate of $\alpha$, serving as a null control without a specific pruning strategy.

We evaluate the performance of models equipped with task vectors derived from the three strategies \textbf{C1}, \textbf{C2}, and \textbf{C3} across a range of pruning rates $\alpha \in \{0.1, 0.2, \dots, 0.9\}$. For \textbf{C3}, considering its inherent randomness, we calculate the average performance over three independent trials. As illustrated in Fig. \ref{fig:valid_exp}, under the condition of \textbf{C1}, model performance across all corresponding tasks remains stable with increasing $\alpha$ and exhibits no significant degradation until a visible decline emerges after $\alpha > 0.7$. Notably, on four tasks including SUN397, Cars, GTSRB, and DTD, performance initially shows a gradual upward trend as $\alpha$ increases and even surpasses the original individual fine-tuned models (Individual) before declining after $\alpha > 0.7$. In contrast, under \textbf{C2}, performance drops sharply from the outset and continues to decrease rapidly as $\alpha$ increases. Furthermore, while the performance decline under \textbf{C3} is more moderate compared to the condition of \textbf{C2}, the inflection point of performance declining appears earlier and the decline is more pronounced than under \textbf{C1}, with a clearly accelerating rate of degradation. 

These comparative observations indicate that a small number of high-magnitude parameters carry the most of task-specific knowledge, demonstrating a typical pulse-activation characteristic. Specifically, only when the parameter magnitude exceeds a certain threshold does it contribute significantly to the target task. Pruning the remaining low-magnitude parameters not only avoids significant performance decline but can even enhance performance while facilitating task vector sparsification. This conclusion confirms that \textbf{task vectors possess an inherent sparse structure that can be effectively leveraged through the simple sparsification strategy presented in C1}. For clarity, we refer to this strategy as P-Spar (Pulse-Sparsification) in the following sections.

\subsubsection{The Quantifiability of Task Vectors}
In this section, we investigate the quantifiability of task vectors. Previous studies have demonstrated that model performance is highly dependent on the qualitative direction of parameter updates, while exhibiting significant robustness to the specific value variations among elements along these directions \cite{ZhuHMD17,LiuXLL0DC24}. Building on this insight, we attempt to further apply binary approximation to the sparsified task vectors. Specifically, we decompose each task vector $\bm{\tau}_k^l$ into three compact components: a binary sparse mask indicating activation positions, a binary sign vector representing parameter polarity, and a scalar scaling factor. This leads to the following approximation form:
\begin{equation}
\widetilde{\bm{\tau}}_{k}^l=\frac{\|\bm{\tau}_{k}^l\odot \mathbf{g}_m(\boldsymbol{\tau}_{k}^l)\|_2}{\|\mathbf{g}_m(\boldsymbol{\tau}_{k}^l)\odot \mathbf{g}_b(\boldsymbol{\tau}_{k}^l) \|_2}* \mathbf{g}_m(\boldsymbol{\tau}_{k}^l)\odot \mathbf{g}_b(\boldsymbol{\tau}_{k}^l),
\label{bin:approx}
\end{equation}
where $\mathbf{g}_b(\bm{\tau}_k^l)$ is the sign vector corresponding to the task vector $\bm{\tau}_k^l$, with its elements defined as
\begin{equation}
    \left(\mathbf{g}_b(\bm{\tau}_k^l)\right)_j=g_{b}({\tau}^l_{k,j})\triangleq\left\{\begin{matrix}
 1, & \text{if } {\tau}^l_{k,j}>0 \\
 -1, & \text{if } {\tau}^l_{k,j}\le 0\\
\end{matrix}\right..
\end{equation}

\begin{figure*}[t]
    \centering
    \includegraphics[width=0.85\textwidth]{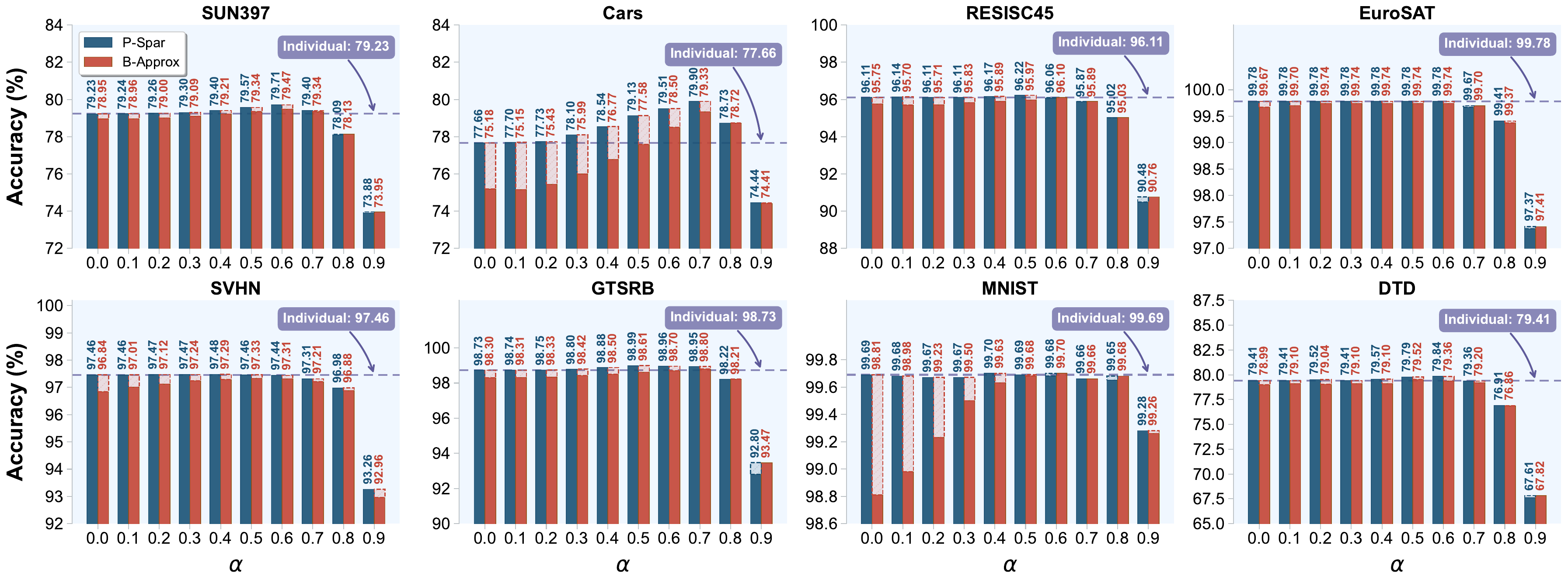}
    \vspace{-10pt}
    \caption{Performance comparison of the ViT-B/32 model equipped with task vectors processed by P-Spar and B-Approx under different pruning rates $\alpha$ across the eight vision tasks. The blue bars represent the task vectors treated solely with P-Spar, while the red bars denote the results of applying further B-Approx. The purple dashed line (Individual) indicates the baseline accuracy of full fine-tuning on each task.}
    \label{fig:valid_quant}
    \vspace{-12pt}
\end{figure*}
The approximation scheme in Eq. \eqref{bin:approx} implies that task vectors can be efficiently encoded using two binary vectors and a floating-point scalar. If such low-bit representations can retain most of the performance of the original task vector, it suggests that task vectors possess an inherent quantizable structure, which can be combined with sparsification to achieve high storage efficiency. To evaluate the feasibility of this approximation (referred to as B-Approx), we apply it to each task vector using the sparse masks generated by P-Spar at various pruning rates $\alpha$. We then test the corresponding performance across the model and datasets described in Section \ref{sec:p-spar}, comparing the results against both the individual fine-tuning baselines and the task vectors sparsified by P-Spar alone.

As illustrated by the experimental results in Fig. \ref{fig:valid_quant}, B-Approx effectively preserves the performance of the task vectors prior to approximation with small accuracy degradation. The largest discrepancy occurs on the Cars dataset at $\alpha=0.1$, where the model performance degrades by $2.55\%$ between the sparsified task vectors before and after B-Approx. More remarkably, as the pruning rate $\alpha$ increases, the performance gap between the task vectors before and after B-Approx exhibits a consistent downward trend across all tasks. This phenomenon is particularly salient on the Cars, GTSRB, and MNIST datasets. Notably, on GTSRB with $\alpha=0.9$, the task vector after B-Approx even outperforms the version using only P-Spar. Furthermore, the performance improvements brought by increased $\alpha$ complement the narrowing gap between pre- and post-approximation task vectors. This synergy enables the quantized task vectors to surpass the original Individual fine-tuning baselines on datasets such as SUN397, Cars, GTSRB, MNIST, and DTD. These findings strongly suggest that task vectors possess an inherent quantizable property. This characteristic can be leveraged by combining quantization with sparsification to construct highly lightweight approximations of task vectors, which ensuring not only strong performance but also achieves a 16-fold reduction in storage overhead, requiring only two binary vectors and a float scalar.

\subsubsection{Dynamic Merging with Binary Task Vectors}
\begin{figure*}[t]
    \centering
    \includegraphics[width=0.8\textwidth]{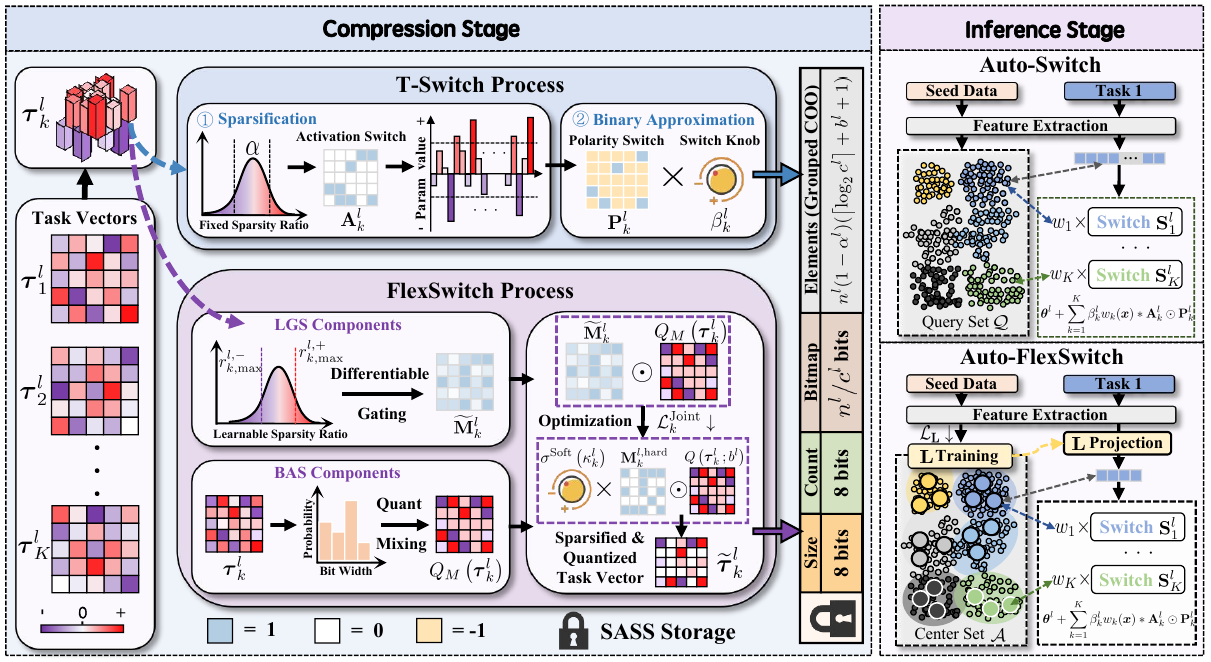}
    \vspace{-7pt}
    \caption{Overview of the proposed methods. The left side illustrates the compression pipelines of T-Switch and FlexSwitch for constructing lightweight task vectors, and the right side illustrates the inference processes of Auto-Switch and Auto-FlexSwitch for dynamic merging.}
	\label{fig:overview} 
    \vspace{-12pt}
\end{figure*}

\paragraph{T-Switch} Inspired by the aforementioned findings, we propose T-Switch, a method for constructing lightweight task vectors as illustrated in Fig. \ref{fig:overview}. By leveraging the intrinsic sparsifiability and quantizability of task vectors, T-Switch enables efficient storage and dynamic application of multi-task capabilities through a series of lightweight 'switches.' Specifically, for the task vector $\bm{\tau}_k^l$ of the $k$-th task in the $l$-th model module, we apply the binarized approximation defined in Eq. \eqref{bin:approx}, deconstructing it into a task switch $\mathbf{S}_k^l = \{\mathbf{A}_k^l, \mathbf{P}_k^l, \beta_k^l\}$ composed of three compact components: 1) \textbf{Activation Switch} $\mathbf{A}_{k}^l = \mathbf{g}_m(\bm{\tau}_k^l) \in \{0, 1\}^{n^l}$, which activates task vector parameters contributing most to $\mathcal{T}_i$; 2) \textbf{Polarity Switch} $\mathbf{P}_k^l = \mathbf{g}_b(\bm{\tau}_k^l) \in \{-1, 1\}^{n^l}$, representing the direction of the task vector for task $\mathcal{T}_i$ at this module; and 3) \textbf{Switch Knob} $\beta_k^l = \frac{\|\bm{\tau}_k^l \odot \mathbf{A}_k^l\|_2}{\|\mathbf{A}_k^l \odot \mathbf{P}_k^l\|_2}$, which provides an approximate scaling of the binarized task vector relative to the full-precision one. With these three components, T-Switch achieves efficient task vector storage while preserving task performance.

\paragraph{Auto-Switch}
Since tasks associated with real-world data may change dynamically, we aim to empower T-Switch with the capabilities of autonomous switching and reconfiguration. While existing dynamic merging strategies often rely on complex, explicitly designed routing modules \cite{lu2024twin,YeHSCHO25}, we introduce Auto-Switch, a training-free dynamic merging mechanism based on inference-time queries. 

We first construct a set of query sets $\mathcal{Q}_k = \{f^{\text{ex}}(\bm{x}; \bm{\Theta}) \mid (\bm{x}, y) \in \mathcal{E}_k \}, k = 1, 2, \dots, K$ using a small portion of exemplar data $\mathcal{E}_k = \{(\bm{x}_i, y_i)\}_{i=1}^N \subset \mathcal{T}_k$ from each task $\mathcal{T}_k$, where $f^{\text{ex}}$ denotes the feature extraction prior to the linear classifier. Note that since the query set only requires input examples, no label information is needed. For each input $\bm{x}$, we perform a KNN search within the global query set $\mathcal{Q} = \cup_{k=1}^K \mathcal{Q}_k$ to search for the set $\mathcal{N}_{\bm{x}}$ consisting of the $C$ nearest neighbors to $f^{\text{ex}}(\bm{x}; \bm{\Theta})$ in $\mathcal{Q}$. This allows us to automatically assign weights to different task switches and perform model merging:
\begin{equation}
\label{eq:auto_merge}
\begin{aligned}
    \hat{\bm{\Theta}}(\bm{x}) &= \mathcal{M} \left( \bm{\Theta}, \left\{\bm{\tau_k}\right\}_{k=1}^K \right) \\
    &= \left\{ \bm{\theta}^l + \sum_{k=1}^K \beta_k^lw_k(\bm{x}) * \mathbf{A}_k^l \odot \mathbf{P}_k^l \right\}_{l=1}^L,
\end{aligned}
\end{equation}
where $w_k(\bm{x}) = \frac{|\mathcal{Q}_k \cap \mathcal{N}_{\bm{x}}|}{|\mathcal{N}_{\bm{x}}|}$ denotes the dynamic weight assigned to the switch of task $\mathcal{T}_k$, and $|\cdot|$ represents the cardinality of the set. By eliminating the need for an explicitly parameterized router, Auto-Switch offers greater flexibility and enables composition without any additional training. Furthermore, since the mechanism requires only a small number of sample feature vectors to be stored rather than raw input data, it further reduces the storage footprint.

\subsection{Adaptive Merging with Learnable Lightweight Task Vectors}
Inspired by the three-component design of T-Switch and the inherent sparsifiability and quantizability of task vectors demonstrated in our previous experiments, we aim to further exploit the potential of these structural properties for more efficient task vector storage. Through an experimental analysis of these components, we observe a distinct heterogeneity across various model modules regarding their optimal sparsity ratios, quantization precision, and magnitude scaling coefficients for different tasks. This suggests that the globally uniform, static configuration of T-Switch still leaves room for efficiency gains through adaptive optimization. Specifically, we identify three key areas for improvement: (1) the non-uniformity of sparse structures, (2) inter-module differences in precision sensitivity, and (3) the suboptimality of the $\ell_2$ magnitude calibration criteria. To address these, we propose the FlexSwitch framework, which shifts the construction of lightweight task vectors from a fixed-rule-based approach to an end-to-end, optimization-driven adaptation. This framework automatically learns the most suitable sparsity rate, bit-width, and scaling factor for each module within every model layer. Furthermore, it dynamically selects the corresponding storage format based on the resulting sparsity patterns and bit-width configurations. Consequently, FlexSwitch maximizes task vector storage efficiency under the constraint of minimal performance degradation. This section elaborates on these aspects in a progressive manner.

\subsubsection{Sensitivity Analysis across Tasks and Modules}
\paragraph{Motivation and Setup}
Considering the functional heterogeneity across different layers and modules of the pre-trained model \cite{DaiDHSCW22}, as well as the unique characteristics of different tasks, the task vectors corresponding to specific modules or layers under varying task contexts are expected to exhibit distinct behaviors in terms of parameter redundancy, quantization sensitivity, and the adaptability of magnitude calibration. To validate this, we design the following sensitivity analysis experiments using CLIP-ViT-B/32 as the base model, following the three-component structure of T-Switch:

\noindent\textbf{Exp I: Sparsity Sensitivity Probing.} We sequentially apply P-Spar with a ratio of $\alpha=0.9$ across two dimensions: \ding{182} Module-type level, targeting one specific module type across all layers at a time; and \ding{183} Layer-wise level, targeting the set of all modules within a single layer. For each probed configuration, we measure the performance drop of the model loaded with the resulted task vectors compared to the original fine-tuned model on each task, while keeping all other modules unchanged. This probing experiment aims to reveal the variation in sparsity sensitivity across different modules and layers under different tasks.

\noindent\textbf{Exp II: Precision Sensitivity Probing.} Following the same two observation dimensions used in \textbf{Exp I}, we apply binarization approximation (i.e., B-Approx with $\alpha=0$) to the task vectors of target modules/layers without any sparsification, and evaluate the resulting performance drop on each task relative to the original fine-tuned model. This experiment is designed to probe the varying sensitivity of various modules and layers to extreme quantization precision.

\noindent\textbf{Exp III: Magnitude Calibration Bias Probing.} In this setup, we apply B-Approx with $\alpha=0$ to all task vectors of all modules across all layers simultaneously. The key difference lies in introducing a tuning factor $\eta \in \{0.1, 0.2, \dots, 2.0\}$ to recalibrate each switch knob individually, i.e., $\beta^l_k\leftarrow\beta^l_k\cdot\eta$, and evaluating the performance drop under each $\eta$ on each task. This experiment aims to investigate the divergent scaling requirements across different tasks and verify whether the static $\ell_2$ norm-based calibration criteria deviate from the optimal scaling magnitude for each task.

\begin{figure*}[t]
    \centering
    \includegraphics[width=0.98\textwidth]{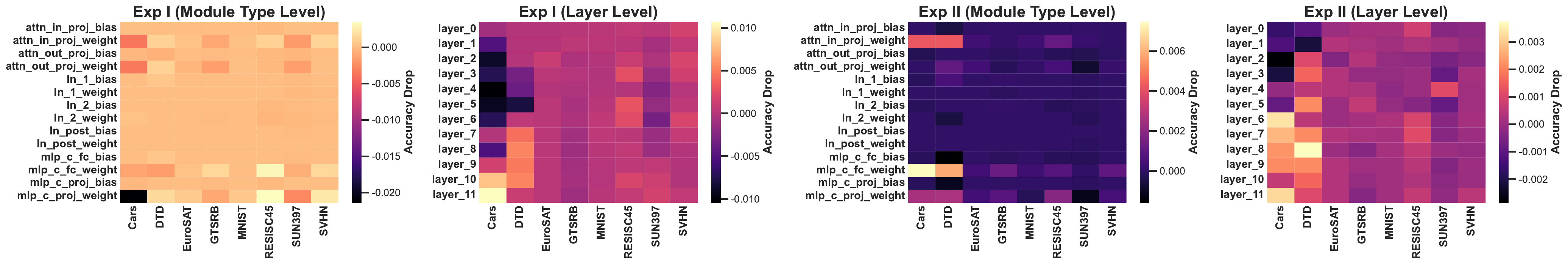}
    \vspace{-7pt}
    \caption{Heatmaps illustrating the sensitivity of different modules/layers to sparsification and quantization. The horizontal axis represents task names, while the vertical axis denotes module types or layer indices. The color indicates the extent of performance degradation (expressed in decimal form, with negative values indicating performance improvements) of the model after applying the corresponding operation to the task vectors of specific modules or layers, compared to the original fine-tuned model. The two plots on the left present the experimental results of \textbf{Exp I}, while the two on the right display the results of \textbf{Exp II}.}
    \label{fig:probe_exp}
    \vspace{-13pt}
\end{figure*}

\begin{figure}[t]
    \centering
    \includegraphics[width=0.42\textwidth]{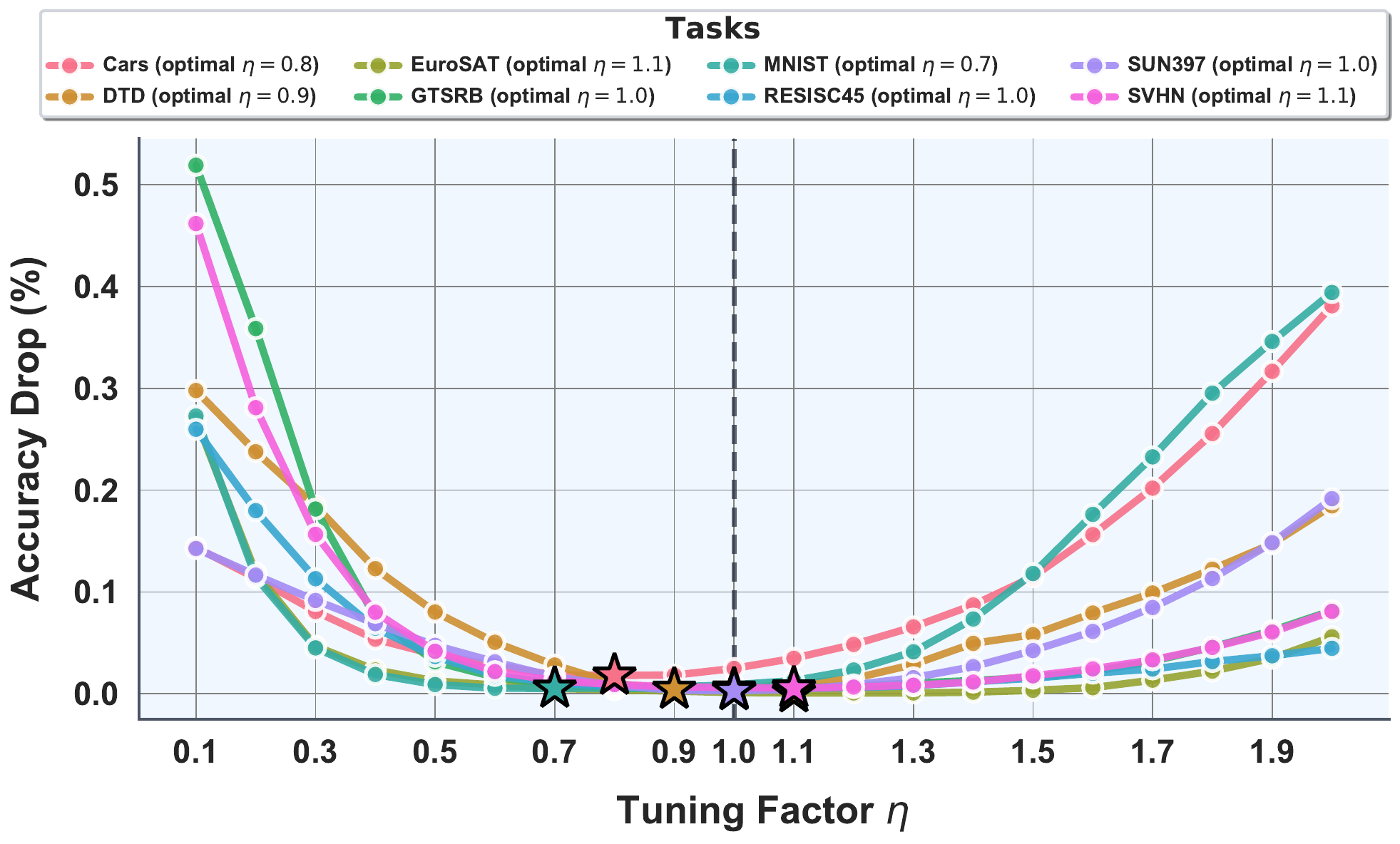}
    \vspace{-8pt}
    \caption{Line charts illustrating the performance degradation across different downstream tasks as the tuning factor $\eta \in [0.1, 2.0]$ varies. Star markers denote task-specific optima (minimum performance drop). The bold vertical dashed line marks $\eta = 1.0$, where no tuning is applied.}
    \label{fig:probe_exp_3}
    \vspace{-12pt}
\end{figure}

\paragraph{Observations and Analysis}
Based on the results of Exp I and Exp II presented in Fig. \ref{fig:probe_exp}, we derive three key observations. 1) At the module level, whether performing sparsification or quantization approximation, the four modules including \texttt{attn\_in\_proj\_weight}, \texttt{attn\_out\_proj\_weight}, \texttt{mlp\_c\_fc\_weight}, and \texttt{mlp\_c\_proj\_weight} exhibit higher sensitivity than other modules. Sparsification or quantization applied to these modules tends to cause more pronounced performance fluctuations. 2) At the layer level, deeper layers demonstrate relatively lower tolerance to sparsification and quantization. Extreme approximations in the deeper stages (Layers 7–11) often result in more pronounced performance degradation, whereas similar operations on shallower layers (Layers 0-5) do not lead to significant performance drops and can even yield some performance improvements. 3) The sensitivity across different modules and layers vary depending on the task. Furthermore, the results of Exp III shown in Fig. \ref{fig:probe_exp_3} reveal that only three tasks (SUN397, GTRSB, and RESISC45) achieve optimal performance with $\eta=1.0$, while other tasks require either further amplification or reduction of the original $\ell_2$ scaling factor to reach their optima, with the optimal values differing across tasks. Synthesizing these observations, we identify three critical bottlenecks limiting the representation efficiency of task vectors:
\begin{enumerate}
\item \textbf{Non-uniformity of sparse structures}: Different modules across layers exhibit varying tolerance to sparsification, making a globally uniform sparsity ratio prone to causing performance degradation in sensitive layers or spatial waste in redundant layers. 
\item \textbf{Inter-module differences in precision sensitivity}: Modules display varying sensitivity to quantization representations, making it challenging to balance storage efficiency and performance across layers/modules with a uniformly extreme quantization bit-width.
\item \textbf{Suboptimality of the $\ell_2$ magnitude calibration criteria}: Static scaling based on $\ell_2$ norm ratios merely pursues approximation in physical magnitude, yet such scaling calibration is often suboptimal. 
\end{enumerate}
Motivated by these insights, we seek to extend T-Switch by transitioning from a construction mode reliant on fixed rules to an adaptive and learnable optimization paradigm, thereby enabling the automated allocation of the most suitable sparsity ratio, quantization bit-width, and scaling factor for the task vector of each module within every layer.

\subsubsection{Learnable Sparsification via Differentiable Gating}
We first introduce the LGS mechanism to enable optimization of sparsity ratios. It assigns a set of learnable positive and negative element thresholds to the task vector $\bm{\tau}_{k}^l$ of each module to construct a soft sparsification gating vector. To ensure that the positive and negative thresholds always remain within the value range of their corresponding signed elements, we first examine the positive and negative elements of the vector and calculate the lower and upper bounds for the absolute values of the positive elements, $[v_{k,\text{min}}^{l,+},v_{k,\text{max}}^{l,+}]$, as well as those for the negative elements, $[v_{k,\text{min}}^{l,-},v_{k,\text{max}}^{l,-}]$. We then assign a set of learnable positive and negative threshold logits $s^{l,+}_{k}\in\mathbb{R}$ and $s^{l,-}_{k}\in\mathbb{R}$, and construct the positive and negative element thresholds $t_{k}^{l,+}$ and $t_{k}^{l,-}$ as follows:
\begin{equation}
\label{eq:6}
\begin{aligned}
t_{k}^{l,+} &= v_{k,\min}^{l,+} + \varphi\left(s^{l,+}_{k}\right) \cdot r_{k,\text{max}}^{l,+}, \quad r_{k,\text{max}}^{l,+} = v_{k,\max}^{l,+} - v_{k,\min}^{l,+}, \\
t_{k}^{l,-} &= v_{k,\min}^{l,-} + \varphi\left(s^{l,-}_{k}\right) \cdot r_{k,\text{max}}^{l,-}, \quad r_{k,\text{max}}^{l,-} = v_{k,\max}^{l,-} - v_{k,\min}^{l,-},
\end{aligned}
\end{equation}
where $\varphi(\cdot): \mathbb{R} \to (0, 1)$ is a normalized smooth mapping function of the form $\varphi(x) = \frac{1}{\pi} \arctan\left(x\right) + 0.5$, which ensures that the parameterization strategy shown in Eq. \eqref{eq:6} strictly constrains the positive and negative thresholds $t_{k}^{l,+}$ and $t_{k}^{l,-}$ within the element range of their corresponding signs. Meanwhile, its gradient decay rate is only $O(1/x^2)$, a property that effectively expands the effective search radius of the threshold parameters $s_{k}^{l,\pm}$ during optimization. Considering that the sparsification process inevitably alters the shape of task vectors, the scaling factor should be coupled with the sparsification strategy to facilitate co-evolution. To this end, we introduce a learnable scaling weight $\kappa_{k}^l$ and combine it with the thresholds to construct a differentiable gating vector $\widetilde{\mathbf{M}}_{k}^{l}$ that unifies sparsity filtering and norm-rescaling as follows:
\begin{equation}
\begin{aligned}
\widetilde{\mathbf{M}}_{k}^{l} &= \sigma^{\text{Soft}}\left(\kappa_{k}^l\right)\cdot \mathbf{M}_{k}^{l}, \\
\mathbf{M}_{k}^{l} &=\left[\sigma^{\text{Sig}}\left(\frac{\bm{\tau}_k^l-\mathbf{1}^l t_{k}^{l,+}}{\rho\cdot r_{k,\text{max}}^{l,+}}\right)+\sigma^{\text{Sig}}\left(\frac{-\mathbf{1}^l t_{k}^{l,-}-\bm{\tau}_k^l}{\rho\cdot r_{k,\text{max}}^{l,-}}\right)\right],
\end{aligned}
\end{equation}
where $\sigma^{\text{Soft}}(\cdot)$ and $\sigma^{\text{Sig}}(\cdot)$ denote the Softplus and Sigmoid functions, respectively. $\mathbf{1}^l$ is an all-ones vector with the same dimension as $\bm{\tau}_k^l$. The temperature parameter $\rho \in (0,1]$, which regulates mask smoothness, is combined with $t_{k}^{l,+}$ and $t_{k}^{l,-}$ to eliminate scaling effects, ensuring its controllability. As $\rho \rightarrow 0$, $\mathbf{M}_{k}^{l}$ approaches an ideal $0$-$1$ mask, where elements within the interval $(-t_{k}^{l,-}, t_{k}^{l,+})$ vanish toward $0$, while those outside converge to $1$. $\sigma^{\text{Soft}}(\kappa_{k}^l)$ serves as a non-negative scaling factor, acting as an adaptive ``switch knob". This gating mechanism provides an evolutionary path from smooth functional approximation to hard-thresholding logic, enabling the joint adaptive optimization of the sparsification strategy and scaling coefficients via backpropagation. To balance the degree of sparsity with performance degradation during optimization, the objective function of LGS is defined as follows:
\begin{equation}
\label{eq:obj_lgs}
\mathcal{L}^{\text{LGS}}_k\left(\mathcal{B}_k\right)= \frac{\sum_{l=1}^L\left\|\mathbf{M}_{k}^{l}\right\|_1}{\sum_{l=1}^Ln_l}+\lambda\mathcal{L}_{k}^{\text{Per}}\left(\mathcal{B}_k\right),
\end{equation}
where the first term promotes structural sparsity by minimizing the magnitudes of elements within each $\mathbf{M}_k^l$, normalized to the range $[0, 1]$ based on the parameter count $n_l$ of each respective module. $\mathcal{B}_k = \{\bm{x}_i\}_{i=1}^B$ denotes a batch of input examples sampled from the $k$-th task. $\mathcal{L}_{k}^{\text{Per}}$ is the performance preservation loss, maintaining predictive consistency between the model with sparsified task vector and the original fine-tuned model, and $\lambda$ is a hyperparameter that controls the intensity of performance preservation. It is well-compatible with current mainstream alignment metrics used for knowledge distillation. Specifically, we consider the following types:

\begin{figure*}[t]
    \centering
    \includegraphics[width=0.85\textwidth]{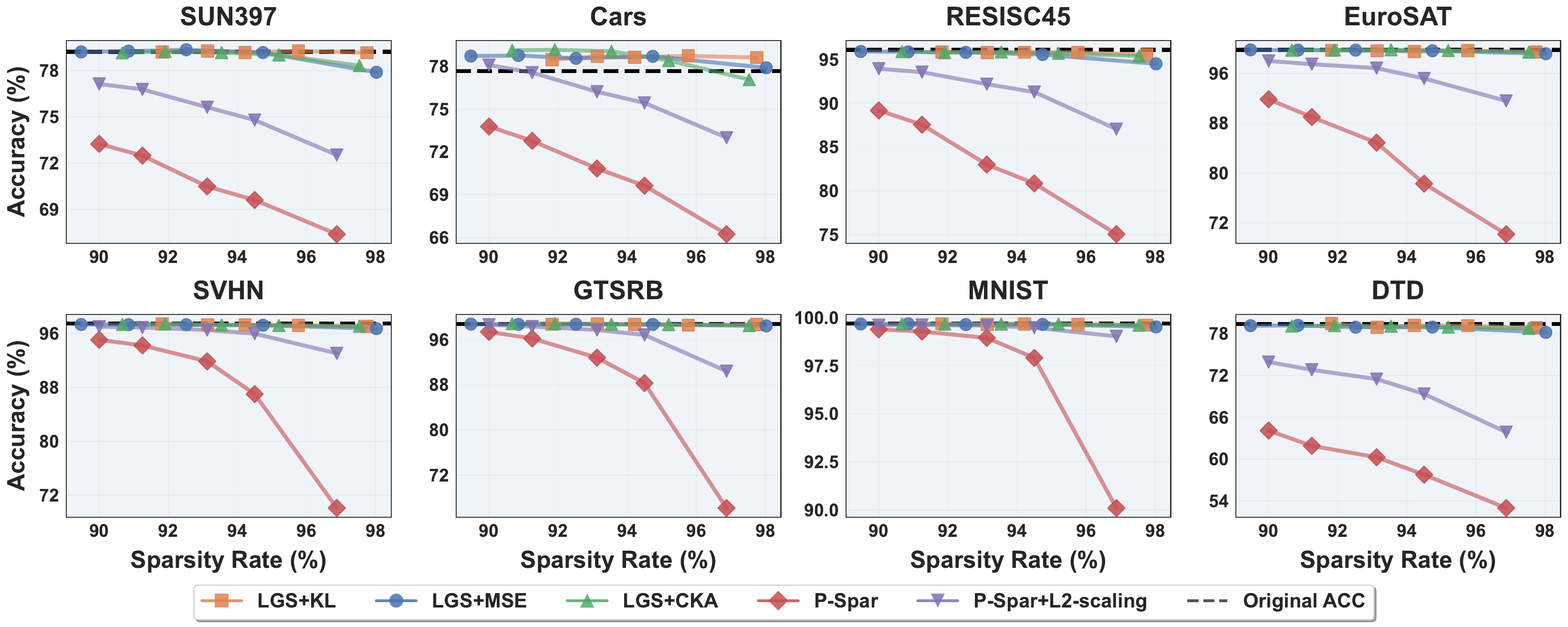}
    \vspace{-8pt}
    \caption{Accuracy of models equipped with task vectors sparsified by LGS with different performance preserving losses and P-Spar across different tasks under different sparsity levels. The dashed line represents the performance of the original fine-tuned model. LGS results are averaged over three trials.}
    \label{fig:valid_LGS}
    \vspace{-12pt}
\end{figure*}
\noindent\textbf{Probability distribution alignment}. This criterion aligns the model's predictive behavior by matching the output probability distributions. As the most widely adopted measure in this context \cite{xu2020knowledge,guo2020online,yang2023online}, we employ Kullback-Leibler (KL) divergence \cite{kullback1951information} as the representative metric. In our setting, its specific form is defined as:
\begin{equation}
\mathcal{L}_{k}^{\text{KL}}\left(\mathcal{B}_k\right)=\frac{T^2}{B}\sum_{i=1}^B\sum_{j=1}^{C_k}q_{i,j}\log\frac{q_{i,j}}{p_{i,j}},
\end{equation}
where $q_{i,j}$ and $p_{i,j}$ denote the $j$-th element of the probability vectors $\mathbf{q}_i$ and $\mathbf{p}_{i}$, respectively. $\mathbf{q}_i = \text{Softmax}(f(\bm{x}_i; \bm{\Theta}_k)/T)$ represents the output probability of the original fine-tuned model for input $\bm{x}_i$, while $\mathbf{p}_i = \text{Softmax}(f(\bm{x}_i; \bm{\Theta}+\hat{\bm{\tau}}_{k})/T)$ corresponds to the model equipped with the soft-sparsified task vector $\hat{\bm{\tau}}_{k}=\left\{\bm{\tau}_{k}^l\odot\widetilde{\mathbf{M}}_{k}^{l}\right\}_{l=1}^L$. Following common practice \cite{cho2019efficacy,zhao2022decoupled}, we set the temperature $T=4$. $C_k$ denotes the dimension of the output logits for the $k$-th task. 

\noindent\textbf{Numerical Metric Alignment}. This criterion aligns model responses by constraining the absolute magnitude of the output values. We employ MSE as the metric, defined as:
\begin{equation}
\mathcal{L}_{k}^{\text{MSE}}\left(\mathcal{B}_k\right) = \frac{1}{B} \sum_{i=1}^B \left\| f(\bm{x}_i; \bm{\Theta}_k) - f(\bm{x}_i; \bm{\Theta}+\hat{\bm{\tau}}_{k}) \right\|_2^2.
\end{equation}

\noindent\textbf{Structural Consistency Alignment}. This criterion ensures that the sparsified model preserves the structured information of the original fine-tuned model within the output space. We employ Centered Kernel Alignment (CKA) to quantify the structural consistency between the outputs of different models:
\begin{equation}
\mathcal{L}_{k}^{\text{CKA}}\left(\mathcal{B}_k\right) = 1 - \frac{\left\| \mathbf{F}_k^\top \mathbf{H} \hat{\mathbf{F}}_k \right\|_F^2}{\left\| \mathbf{F}_k^\top \mathbf{H} \mathbf{F}_k \right\|_F \left\| \hat{\mathbf{F}}_k^\top \mathbf{H} \hat{\mathbf{F}}_k \right\|_F},
\end{equation}
where $\mathbf{F}_k = [f(\bm{x}_1; \bm{\Theta}_k), \dots, f(\bm{x}_B; \bm{\Theta}_k)]^\top \in \mathbb{R}^{B \times C_k}$ and $\hat{\mathbf{F}}_k = [f(\bm{x}_1; \bm{\Theta}+\hat{\bm{\tau}}_{k}), \dots, f(\bm{x}_B; \bm{\Theta}+\hat{\bm{\tau}}_{k})]^\top \in \mathbb{R}^{B \times C_k}$ denote the output matrices of the original model and the model equipped with the soft sparsified task vector, respectively, on the input batch $\mathcal{B}_k$. $\mathbf{H} = \mathbf{I} - \frac{1}{B}\mathbf{1}\mathbf{1}^\top$ is the centering matrix.

For the LGS training process, we randomly select $N=100$ examples per task as the example set. The batch size $B$ is set to 32, and the optimization is performed for 500 steps. To ensure smooth convergence of the threshold parameters, we initialize the temperature parameter $\rho=1.0$ and apply an exponential decay of $\rho \leftarrow 0.9 \cdot \rho$ every 10 optimization steps. Upon completion, the final hard-sparsified gating mask is computed as $(\mathbf{M}_{k, \text{hard}}^{l})_{i,j} = \mathbbm{1}((\mathbf{M}_{k}^{l})_{i,j} > 0.5)$, which determines the final sparsified task vector $\hat{\bm{\tau}}_{k}=\left\{\sigma^{\text{Soft}}\left(\kappa_{k}^l\right)\cdot\bm{\tau}_{k}^l\odot\mathbf{M}_{k, \text{hard}}^{l}\right\}_{l=1}^L$.
\begin{table}[h]
\vspace{-10pt}
\centering
\caption{Hyperparameter configurations of $\lambda$ for different performance preservation losses.}
\renewcommand{\arraystretch}{0.7}
\label{tab:1}
\begin{adjustbox}{width=0.37\textwidth}
\begin{tabular}{lc} 
\toprule
\textbf{Performance Preservation Loss} & \textbf{$\lambda$ Settings} \\ \midrule
KL & \{0.1, 0.3, 0.5, 0.7, 0.9\} \\ \addlinespace[0.5em]
MSE           & \{0.01, 0.05, 0.09, 0.13, 0.17\} \\ \addlinespace[0.5em]
CKA           & \{1.0, 3.0, 5.0, 7.0, 9.0\} \\ \bottomrule
\end{tabular}
\end{adjustbox}
\end{table}
To validate the effectiveness of the proposed LGS in balancing sparsity and performance, we compare it with P-Spar in the verification experiments described in Sec. \ref{sec:p-spar}. We follow Tab. \ref{tab:1} to configure the hyperparameter for the performance preservation loss, selecting specific $\lambda$ values for different types of loss functions to ensure that each approach achieves a comparable sparsity level. Meanwhile, the pruning rate $\alpha$ for P-Spar on each task is set according to the sparsity level achieved by LGS under different hyperparameter choices on the corresponding task. To account for the effect of the scaling weight $\kappa_{k}^l$, we introduce a controlled variant where the L2 scaling applied in T-Switch is incorporated into the task vectors after P-Spar, restoring them to their original norm. Experiments of LGS are run three times independently to eliminate the impact of randomness. As shown in Fig. \ref{fig:valid_LGS}, LGS effectively preserves or even surpasses (on the Cars dataset) the performance of the fine-tuned model regardless of the performance preserving loss employed. Even at high sparsity levels approaching $98\%$, LGS maintains the model's performance well. In contrast, P-Spar exhibits rapid performance degradation at sparsity levels above $90\%$. The performance drop remains significant even with L2 scaling applied to adjust the magnitude. Notably, at a sparsity level of $97\%$, P-Spar+L2-Scaling results in an accuracy drop of over 10 points on DTD. This comparison strongly demonstrates the effectiveness of LGS in balancing sparsity and model performance.

\subsubsection{Bitwidth Allocation Guided by Task Utility} 
Building upon the sparse patterns learned by LGS, we introduce BAS to transition from fixed binarization to fine-grained adaptive bit-width allocation. The BAS mechanism selects quantization bit-widths from a broader set $\mathcal{W}=\{1, 2, 4, 8\}$. To achieve this, it first assigns a set of learnable bit-width distribution parameters $\mathbf{w}^l\in\mathbb{R}^4$ to the task vector $\bm{\tau}_k^l$ of each module, and then constructs bit-width distribution probabilities to implement a differentiable bit-width mixing strategy:
\begin{equation}
Q_{M}\left(\bm{\tau}_k^l\right)=\sum_{i=1}^{\left|\mathcal{W}\right|}\frac{\exp\left(w_i^l/\omega\right)}{\sum_{j=1}^{\left|\mathcal{W}\right|}\exp\left(w_j^l/\omega\right)}\cdot Q\left(\bm{\tau}_k^l\,;\mathcal{W}_i\right),
\end{equation}
here, $w_i^l$ is the $i$-th component of $\mathbf{w}^l$, $\mathcal{W}_i$ is the $i$-th element of $\mathcal{W}$, and $\omega$ is the temperature parameter controlling the smoothness of the bit-width distribution. As $\omega \to 0$, the distribution converges to a one-hot form, effectively approaching a discrete, single bit-width selection. $Q(\bm{\tau}_k^l; b)$ represents the asymmetric quantization operation with bit-width $b$:
\begin{equation}
\left(Q\left(\bm{\tau}_k^l\,;b\right)\right)_j = -v_{k,\text{max}}^{l,-}+\left(\mathcal{I}\left({\tau}_{k,j}^l\,;b\right)+0.5\right)\cdot\delta^l,
\end{equation}
where $\delta^l=\frac{v_{k,\text{max}}^{l,+}+v_{k,\text{max}}^{l,-}}{2^b}$ is the quantization step size, and $\mathcal{I}\left({\tau}_{k,j}^l\,;b\right) = \text{Clamp}\left(\left \lfloor \frac{{\tau}_{k,j}^l+ v_{k,\text{max}}^{l,-}}{\delta^l} \right\rfloor,1, 2^b\right)-1$ represents the quantization bin index for the element ${\tau}_{k,j}^l$.

Building upon the aforementioned bit-width mixing strategy, BAS incorporates a bit-width optimization regularization term into Eq. \eqref{eq:obj_lgs}, thereby combining the optimization of quantization and sparsification strategies to construct the following joint optimization objective:
\begin{equation}
\label{eq:joint_opt}
\mathcal{L}^{\text{Joint}}_k\left(\mathcal{B}_k\right)= \frac{\sum_{l=1}^L\left\|\mathbf{M}_{k}^{l}\right\|_1}{\sum_{l=1}^Ln_l}+\frac{\sum_{l=1}^L \overline{w}^l}{L\max\left(\mathcal{W}\right)}+\lambda\mathcal{L}_{k}^{\text{Per}}\left(\mathcal{B}_k\right),
\end{equation}
where $\overline{w}^l=\sum_{i=1}^{\left|\mathcal{W}\right|}\frac{\exp\left(w_i^l/\omega\right)}{\sum_{j=1}^{\left|\mathcal{W}\right|}\exp\left(w_j^l/\omega\right)}\cdot \mathcal{W}_i$ represents the mixed bit-width for the $l$-th module, and $\max\left(\mathcal{W}\right)$ denotes the maximum bit-width in $\mathcal{W}$, which normalizes the regularization term to the $[0, 1]$ range, thereby aligning its scale with the sparsity objective. Note that the performance-preservation loss in this optimization process is computed using the task vector $\widetilde{\bm{\tau}}_{k}=\left\{\widetilde{\mathbf{M}}_{k}^{l}\odot Q_{M}\left(\bm{\tau}_k^l\right)\right\}_{l=1}^L$, which accounts for the combined effect of soft sparsification and mixed bit-width quantization.
Upon completion of the optimization, the final bit-width $b^l$ for the $l$-th module is selected as the one corresponding to the maximum bit-width distribution parameter:
\begin{equation}
b^l = \mathcal{W}_{\hat{i}}, \quad \hat{i} = \mathop{\arg\max}_{i \in \{1,\dots, \left|\mathcal{W}\right|\}} w_i^l,
\end{equation}
this yields the final sparsified and quantized task vector $\widetilde{\bm{\tau}}_{k}=\left\{\sigma^{\text{Soft}}\left(\kappa_{k}^l\right)\cdot\mathbf{M}_{k}^{l, \text{hard}}\odot Q\left(\bm{\tau}_k^l\,;b^l\right)\right\}_{l=1}^L$.

\begin{figure}[t]
    \centering
    \includegraphics[width=0.48\textwidth]{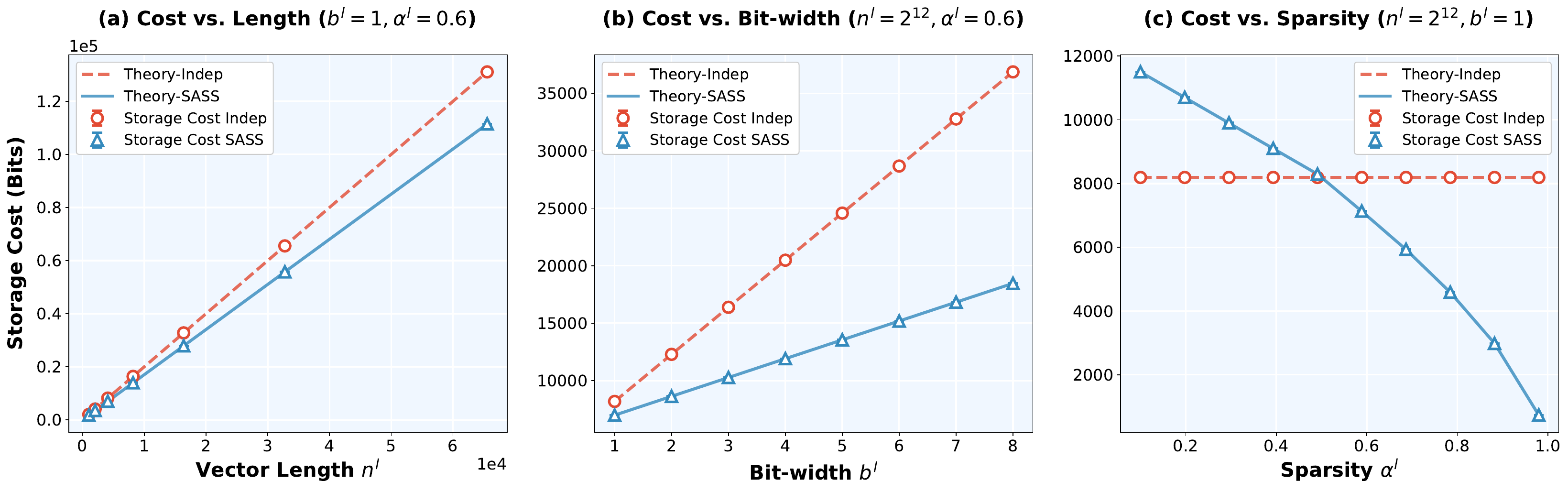}
    \vspace{-7pt}
    \caption{Storage comparison between the SASS and Indep schemes under different parameter configurations. The curves represent the theoretical number of storage bits, while the scattered points indicate the average actual storage bits over 10 independent trials, with error bars denoting the standard deviation.}
    \label{fig:valid_SASS}
    \vspace{-12pt}
\end{figure}

\subsubsection{Cost-Aware Storage Format Selection}
\label{sec:sass}
For compressed task vectors, a straightforward storage scheme is to independently store the binary mask and the quantized values (referred to as Indep). This approach allocates a fixed bit budget (mask + quantization bits) to every element, treating zero and non-zero entries identically. Consequently, it fails to fully exploit the inherent sparsity of the task vectors. To unlock the potential of the high sparsity ratios achieved by LGS, we introduce SASS, which leverages a grouped COO storage format to adaptively select the optimal group count. Concretely, for a task vector of length $n^l$, it is first evenly partitioned into $\frac{n^l}{c^l}$ groups of size $c^l$ ($c^l \mid n^l$). A bitmap is then constructed to indicate the positions of non-zero groups, where a group containing at least one non-zero element is marked as $1$, and an all-zero group as $0$, resulting in a storage overhead of $n^l/c^l$ bits. Subsequently, for each non-zero group, we store the indices of non-zero elements (using $\left\lceil \log_2 c^l \right\rceil$ bits each) alongside their quantized values. Assuming that the occurrences of non-zero elements are i.i.d. following a Bernoulli distribution with an expected sparsity ratio $\alpha^l$ and a quantization bit-width $b^l$, the expected storage overhead for this scheme is:
\begin{equation}
\label{eq:storage}
\text{Storage}^{\textbf{SASS}} = 35+\frac{n^l}{c^l} + n^l(1-\alpha^l)(\left\lceil \log_2 c^l \right\rceil + b^l+1)\quad\text{bits},
\end{equation}
where the constant $35$ accounts for the storage of the group count and the group size. We consider a maximum group size of $256$, for which $8$ bits are allocated. Since the maximum vector length $n^l$ under consideration is $2^{27}$, which covers the maximum weight vector dimensions of mainstream models with 7B parameters or fewer, $27$ bits are reserved to store the group count information. Within the parenthetical term $(\lceil \log_2 c^l \rceil + b^l + 1)$, in addition to storing the intra-group index ($\lceil \log_2 c^l \rceil$ bits) and the quantized value ($b^l$ bits) for each non-zero element, we introduce an extra 1-bit flag to signify the end of the current group, which ensures deterministic decoding of the stored data driven by the bitmap. Note that, when ignoring the ceiling function, the above expression possesses a unique minimum point at $c^l = \frac{\ln 2}{1-\alpha^l}$ that depends solely on the sparsity ratio. Consequently, we adaptively select $c^l$ within the feasible domain $\mathcal{C}$ according to the following rule:
\begin{equation}
\label{eq:c_select}
c^l = \arg\min_{c\in\mathcal{C}}\left|\frac{\ln 2}{(1-\alpha^l)}-c\right|,\,\, \mathcal{C}= \left\{c\mid c\in\mathbb{Z}^+, c\mid n^l\right\}.
\end{equation}

\begin{figure*}[t]
    \centering
    \includegraphics[width=0.85\textwidth]{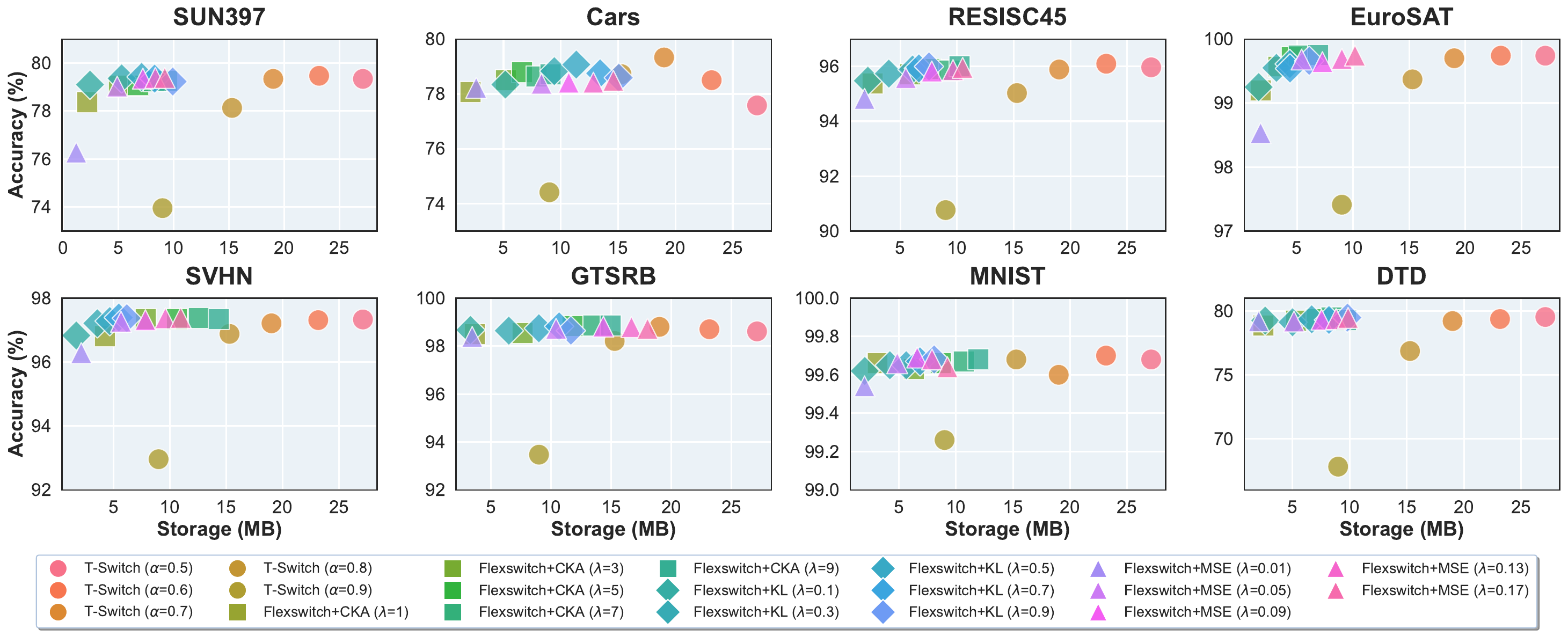}
    \vspace{-7pt}
    \caption{Comparison of performance and storage overhead (MB) between FlexSwitch and T-Switch across different tasks, using SASS as the unified storage scheme. All results report the mean values from three independent runs.}
    \label{fig:valid_FlexSwitch}
    \vspace{-12pt}
\end{figure*}

We further examine the theoretical storage overhead difference $\text{Storage}^{\text{SASS}} - \text{Storage}^{\text{Indep}}$ between SASS and the independent storage of mask vectors and quantized task vectors (Indep), where $\text{Storage}^{\textbf{Indep}}=(b^l+1)n^l$. Also under the condition that ignores the ceiling operations in $\text{Storage}^{\text{SASS}}$, when the optimal group size $c^l = \frac{\ln 2}{1-\alpha^l}$ is substituted, this difference function decreases with respect to both $b^l$ and $\alpha^l$; furthermore, it also decreases with $n^l$ provided that $\alpha > 0.5$. This implies that SASS yields a more pronounced storage advantage over the Indep scheme at higher sparsity levels and larger quantization bit-widths. Moreover, increased sparsity further amplifies SASS's storage benefits as the parameter scale of the task vectors grows. To verify this theoretical expectation and verify the accuracy of the proposed theoretical storage occupancy formula, we compare the theoretical bit counts and actual storage bits of both schemes by varying $n^l, b^l$, and $\alpha^l$, with the group size selected according to Eq. \eqref{eq:c_select}. For the actual storage tests, we generate sparse task vectors with sparsity rate $\alpha$ by constructing mask vectors according to a Bernoulli distribution with mean $1-\alpha$. The parameter ranges were set to $n^l \in \{2^{10}, 2^{11}, \dots, 2^{16}\}$ and $b^l \in \{1, 2, \dots, 8\}$, while 10 observation points for $\alpha^l$ were sampled uniformly within the range $[0.1, 0.98]$. Each configuration was run $10$ times to plot the mean values and standard deviations. Results shown in Fig. \ref{fig:valid_SASS} indicate that SASS consistently maintains lower storage costs than the Indep scheme across all parameter configurations when $\alpha=0.6$ (satisfying the $\alpha > 0.5$ theoretical condition). Meanwhile, the actual storage bits at the sampling points perfectly match the theoretical curves. As $n^l$ and $b^l$ increase, the storage advantage of SASS over Indep grows consistently. Notably, when $\alpha > 0.5$, SASS begins to yield a positive storage gain that expands rapidly, thereby validating our theoretical expectations. It is worth mentioning that at an sparsity of $\alpha^l = 0.98$, the storage overhead of SASS is less than $1/8$ of that required by Indep, confirming the efficacy of SASS in leveraging the high sparsity of task vectors to optimize storage efficiency.

\subsubsection{Auto-FlexSwitch}
Based on the joint optimization of LGS and BAS, combined with the efficient storage strategy of SASS, we construct the complete FlexSwitch framework. As illustrated in Fig. \ref{fig:overview}, for each task vector $\bm{\tau}_k$ to be sparsified, the framework first jointly optimizes its sparsification and quantization strategies based on Eq. \eqref{eq:joint_opt}, synergistically improving compression efficiency while preserving task performance. Subsequently, SASS is employed to store the compressed task vector efficiently, directly translating the benefits of high sparsity rates and low bit-widths achieved by LGS and BAS into actual savings in disk space.

To verify the efficiency advantage of FlexSwitch over T-Switch, we adopt SASS as the unified storage scheme and conduct a comparative evaluation of storage efficiency and performance between FlexSwitch with different performance retention losses and T-Switch with varying sparsity rates $\alpha \in \{0.5, 0.6, \dots, 0.9\}$. For FlexSwitch, the performance retention loss hyperparameter $\lambda$ is aligned with Tab. \ref{tab:1}. The temperature parameters $\rho$ and $\omega$, which regulate sparse gating and bit-width distribution smoothness, are both initialized to $1.0$ and decay exponentially by a factor of $0.9$ every 10 optimization steps. The learning rates for LGS and BAS parameters are set to $0.05$ and $0.1$, respectively. Optimization is performed for $500$ iterations using the Adam optimizer on an exemplar set consisting of $N=100$ randomly selected samples per task.

\begin{table*}[h]
\centering
\caption{Hyperparameter settings for different methods across various model architectures. “--” indicates that the corresponding parameter is not applicable.}
\label{tab:2}
\renewcommand{\arraystretch}{0.6}
\vspace{-5pt}
\resizebox{0.85\textwidth}{!}{
\begin{tabular}{lccccccccc}
\toprule
\multirow{2}{*}{Methods} & \multicolumn{3}{c}{ViT-B/32} & \multicolumn{3}{c}{ConvNeXt} & \multicolumn{3}{c}{ViT-L/14} \\
\cmidrule(lr){2-4} \cmidrule(lr){5-7} \cmidrule(lr){8-10}
& Sparsity Rate & Sample Size & Scaling Coef & Sparsity Rate & Sample Size & Scaling Coef & Sparsity Rate & Sample Size & Scaling Coef \\
\midrule
Task-Arithmetic & -- & -- & 0.2 & -- & -- & 0.2 & -- & -- & 0.3 \\
DARE & 0.4 & -- & 0.2 & 0.1 & -- & 0.2 & 0.6 & -- & 0.4 \\
TIES-Merging & 0.8 & -- & 0.8 & 0.6 & -- & 0.8 & 0.8 & -- & 0.8 \\
Fisher Merging & -- & 8192 & -- & -- & 8192 & -- & -- & 8192 & -- \\
DF-Merge & -- & 14400 & -- & -- & 14400 & -- & -- & 14400 & -- \\
RegMean & -- & 12800 & -- & -- & 12800 & -- & -- & 12800 & -- \\
AdaMerging & -- & 64000 & -- & -- & 64000 & -- & -- & 64000 & -- \\
AdaMerging++ & 0.5 & 64000 & -- & 0.7 & 64000 & -- & 0.3 & 64000 & -- \\
Twin-Merging & -- & 8000 & 0.3 & -- & 8000 & 0.3 & -- & 8000 & 0.2 \\
MoW-Merging & -- & 6824 & -- & -- & 6824 & -- & -- & 6824 & -- \\
Auto-Switch & 0.7 & 800 & -- & 0.5 & 800 & -- & 0.6 & 800 & -- \\
Auto-FlexSwitch & -- & 800 & -- & -- & 800 & -- & -- & 800 & -- \\
\bottomrule
\end{tabular}}
\vspace{-10pt}
\end{table*}

\begin{table*}[t]
\centering
\caption{Performance comparison on ViT-B/32 across eight visual classification tasks. The best and second-best results among all compared merging methods are highlighted in \textbf{bold} and \underline{underlined}, respectively.}
\vspace{-5pt}
\label{tab:3}
\renewcommand{\arraystretch}{0.7}
\resizebox{0.85\textwidth}{!}{%
\begin{tabular}{c l c c c c c c c c c c}
\toprule
\multirow{2}{*}{Type} & \multirow{2}{*}{Method} & \multirow{2}{*}{Storage (MB)} & \multicolumn{9}{c}{Accuracy (\%)} \\
\cmidrule(lr){4-12}
 & & & SUN397 & Cars & RESISC45 & EuroSAT & SVHN & GTSRB & MNIST & DTD & AVG \\
\midrule
\multirow{3}{*}{--} & Pre-trained & -- & 62.31 & 59.64 & 60.22 & 45.81 & 31.61 & 32.56 & 48.25 & 44.41 & 48.10 \\
 & Traditional MTL & -- & 73.20 & 74.76 & 93.43 & 98.89 & 96.67 & 97.28 & 99.40 & 76.60 & 88.78 \\
 & Individual & -- & 79.23 & 77.66 & 96.11 & 99.78 & 97.46 & 98.73 & 99.69 & 79.41 & 91.01 \\
\midrule
\multirow{11}{*}{\rotatebox{90}{Static}} & Weight-Averaging \cite{WortsmanIGRLMNF22} & -- & 64.72 & 63.34 & 71.44 & 72.74 & 64.16 & 52.80 & 87.46 & 50.11 & 65.85 \\
 & Task-Arithmetic \cite{IlharcoRWSHF23} & -- & 63.49 & 62.06 & 72.00 & 78.59 & 74.42 & 65.08 & 94.00 & 52.18 & 70.23 \\
 & DARE \cite{Yu0Y0L24} & -- & 63.51 & 62.06 & 72.08 & 78.37 & 74.62 & 65.13 & 94.07 & 52.34 & 70.27 \\
 & Ties-Merging \cite{YadavTCRB23} & -- & 62.74 & 61.42 & 73.25 & 85.70 & 80.67 & 73.19 & 97.44 & 54.89 & 73.66 \\
 & Fisher Merging \cite{MatenaR22} & -- & 66.16 & 67.07 & 72.41 & 66.89 & 81.44 & 66.09 & 85.35 & 51.81 & 69.65 \\
 & DF-Merge \cite{LeeLWWCW25} & -- & 61.75 & 64.53 & 74.02 & 61.22 & 95.37 & 77.71 & 90.34 & 62.93 & 73.48 \\
 & RegMean \cite{Jin0P023} & -- & 67.86 & 66.87 & 82.37 & 92.96 & 86.41 & 81.95 & 97.25 & 63.19 & 79.86 \\
 & AdaMerging \cite{YangW00G0T24} & -- & 64.17 & 68.85 & 78.86 & 93.19 & 85.90 & 91.34 & 97.36 & 60.00 & 79.96 \\
 & AdaMerging++ \cite{YangW00G0T24} & -- & 65.74 & 69.89 & 81.14 & 94.11 & 86.89 & 90.16 & 98.11 & 61.60 & 80.96 \\
\midrule
\multirow{8}{*}{\rotatebox{90}{Dynamic}} & EMR-Merging \cite{huang2024emr} & 540.19 & 75.20 & 72.75 & 93.49 & 99.52 & 96.86 & 98.12 & 99.58 & 74.36 & 88.74 \\
 & Twin-Merging \cite{lu2024twin} & 711.10 & 74.88 & 73.26 & 95.13 & 99.67 & \underline{97.36} & 98.41 & 99.65 & 78.56 & 89.62 \\
 & MoW-Merging \cite{YeHSCHO25} & 2171.89 & 68.05 & 68.95 & 93.59 & 99.56 & 97.21 & 98.63 & 99.66 & 72.61 & 87.28 \\
 & Auto-Switch & 185.28 & 77.45 & \underline{78.47} & 94.48 & \textbf{99.74} & 97.31 & \underline{98.68} & \textbf{99.70} & 78.19 & 90.51 \\
 & Auto-FlexSwitch+CKA ($\lambda=3$) & \underline{47.26} & \underline{79.11} & \textbf{79.12} & \textbf{95.92} & 99.59 & \textbf{97.45} & 98.61 & 99.65 & \textbf{79.79} & \textbf{91.16} \\
 & Auto-FlexSwitch+KL ($\lambda=0.3$) & \textbf{41.51} & \textbf{79.41} & 78.40 & \underline{95.71} & 99.67 & 97.11 & \textbf{98.70} & \underline{99.67} & 79.36 & \underline{91.00} \\
 & Auto-FlexSwitch+MSE ($\lambda=0.05$) & 50.72 & 79.03 & 78.21 & 95.57 & \underline{99.70} & 97.33 & 98.65 & 99.64 & \underline{79.41} & 90.94 \\
\bottomrule
\end{tabular}%
}
\vspace{-10pt}
\end{table*}

As shown in Fig. \ref{fig:valid_FlexSwitch}, regardless of the combined performance retention loss, FlexSwitch consistently exhibits a significant Pareto advantage. Specifically, it achieves performance comparable to T-Switch while consuming considerably less storage than the latter's substantially higher overhead. In contrast, when the sparsity rate $\alpha$ of T-Switch increases to $0.9$, it no longer maintains task performance effectively, exhibiting clear accuracy degradation. Notably, FlexSwitch achieves performance on par with T-Switch (which requires $\sim 20$ MB or more) using only about $5$ MB of storage across all tasks. This strongly demonstrates that through the synergistic operation of LGS and BAS, FlexSwitch can more precisely eliminate redundancy in task vectors, enabling an efficient trade off between performance and storage efficiency.

Furthermore, as the original feature space of pre-trained models is typically optimized for general representations, feature distributions across multiple tasks may overlap in scenarios with high semantic similarity. This overlap limits the retrieval accuracy of Auto-Switch, which relies on the Euclidean distance of raw features for KNN search. To enhance the discriminability and efficiency of task queries, we introduce a KNN-based merging mechanism integrated with a learnable low-rank metric. Formally, given the constructed query sets $\left\{\mathcal{Q}_k\right\}_{k=1}^K$, we first perform K-Means clustering within each $\mathcal{Q}_k$ to extract $E$ representative centers, denoted as $\bm{\mu}_{i}^k$ for $1 \le i \le E$, to form a condensed reference set $\mathcal{A}^k = \{\bm{\mu}_i^k\}_{i=1}^{E}$. Subsequently, we learn a low-rank projection matrix $\mathbf{L} \in \mathbb{R}^{r \times e}$, where $e$ denotes the output dimension of the pre-trained backbone $f^{\text{ex}}(\cdot; \bm{\Theta})$ and $r \ll e$ is the projection rank. Consequently, the distance between the feature output of an input $\bm{x}$ and a reference center $\bm{\mu}$ is measured by the following metric $d_{\mathbf{L}}\left(\cdot,\cdot\right) \rightarrow \mathbb{R}^+$:
\begin{equation}
d_{\mathbf{L}}\left(f^{\text{ex}}(\bm{x}; \bm{\Theta}),f^{\text{ex}}(\bm{x}'; \bm{\Theta})\right) = \left\|\mathbf{L}f^{\text{ex}}(\bm{x}; \bm{\Theta})- \mathbf{L}f^{\text{ex}}(\bm{x}'; \bm{\Theta})\right\|_2
\end{equation}
to identify the neighborhood set $\mathcal{N}_{\bm{x}}$. We then optimize $\mathbf{L}$ by minimizing the following objective over query sets $\bigcup_{k=1}^{K} \mathcal{A}_k$:
\begin{equation}
\mathcal{L}_{\mathbf{L}} = -\frac{1}{KN}\sum_{i=1}^{KN} \log \left( \frac{\sum_{\bm{z} \in{ \mathcal{N}_{\bm{x}_i} \cap \mathcal{A}_{k_i}}} (d_{\mathbf{L}}(f^{\text{ex}}(\bm{x}_i; \bm{\Theta}), \bm{z}))^{-1}}{\sum_{\bm{z} \in \mathcal{N}_{\bm{x}_i}} (d_{\mathbf{L}}(f^{\text{ex}}(\bm{x}_i; \bm{\Theta}), \bm{z}))^{-1}} \right).
\end{equation}
where $k_i$ denotes the task identity of the input $\bm{x}_i$. Driven by this objective, the low-rank projection matrix $\mathbf{L}$ adaptively reconfigures the original feature distribution. Once optimized, this metric is employed to perform the retrieval-based merging as defined in Eq. \eqref{eq:auto_merge}. By reinforcing task-relevant discriminative dimensions within a low-dimensional space, this mechanism further ensures accurate task vector retrieval with negligible storage overhead. Integrating this inference-time merging mechanism with the lightweight task vectors constructed by FlexSwitch yields Auto-FlexSwitch, the overall pipeline of which is illustrated in Fig. \ref{fig:overview}.

\section{Experiments}
In this section, we evaluate the effectiveness of the proposed methods through extensive experiments, where we compare our approach with various baselines across different downstream tasks and model architectures to demonstrate its superior performance and storage efficiency. Furthermore, we conduct systematic ablations on individual components of the proposed methods. Additionally, to explore the application boundaries and generalizability of our methods, we validate the potential of FlexSwitch itself as a lightweight storage scheme for fine-tuned LLM weights. All experiments are conducted on multiple NVIDIA H200 GPUs.

\subsection{Merging Experiments on Diverse Downstream Scenarios}
\label{sec:main_results}
To verify the generalizability and effectiveness of the proposed methods across diverse tasks, we compare their performance and storage efficiency against multiple baselines in three different downstream scenarios, including image classification (Sec. \ref{sec:img_classify}), object detection (Sec. \ref{sec:img_obj}), and natural language understanding (Sec. \ref{sec:language_und}).

\subsubsection{Merging on Image Classification Tasks}
\label{sec:img_classify}

\paragraph{Experimental Settings}
Following common practice in model merging \cite{huang2024emr,lu2024twin,YeHSCHO25}, we evaluate our methods on eight visual classification datasets: SUN397 \cite{XiaoHEOT10}, Cars \cite{Krause0DF13}, RESISC45 \cite{ChengHL17}, EuroSAT \cite{HelberBDB19}, SVHN \cite{netzer2011reading}, GTSRB \cite{StallkampSSI11}, MNIST \cite{Deng12}, and DTD \cite{CimpoiMKMV14}. To thoroughly verify the effectiveness of our methods, we select three backbones with diverse architectural characteristics and parameter scales, including the Transformer-based CLIP visual encoders, ViT-B/32 \cite{RadfordKHRGASAM21} and the larger-scale ViT-L/14, alongside the convolutional-based ConvNeXt \cite{0003MWFDX22}. To quantify the performance and efficiency of different methods, we evaluate both task accuracy (\%) and the additional parameter storage overhead (MB) introduced by each dynamic merging approach. For all methods involving randomness, we report the average results across three independent runs.

\paragraph{Baselines}
We compare our methods with various classical and SOTA baselines for both static and dynamic model merging, including 1) Weight-Averaging \cite{WortsmanIGRLMNF22}, 2) Task-Arithmetic \cite{IlharcoRWSHF23}, 3) DARE \cite{Yu0Y0L24}, 4) TIES-Merging \cite{YadavTCRB23}, 5) Fisher Merging \cite{MatenaR22}, 6) DF-Merge \cite{LeeLWWCW25}, 7) RegMean \cite{Jin0P023}, 8) AdaMerging \cite{YangW00G0T24}, 9) AdaMerging++ \cite{YangW00G0T24}, 10) EMR-Merging \cite{huang2024emr}, 11) Twin-Merging \cite{lu2024twin}, and 12) MoW-Merging \cite{YeHSCHO25}. For baselines involving exemplar samples, we strictly follow the sample sizes recommended in their original papers. For baselines that involve scaling coefficients or sparsity rates, we conduct an grid search on each method to determine their optimal parameter configurations. The resulting parameter settings are listed in Tab. \ref{tab:2}. For Auto-FlexSwitch, the learning rates for LGS and BAS are set to 5e-2 and 1e-1, respectively. Optimization is conducted using the Adam optimizer for 500 steps, with temperature parameters $\rho$ and $\omega$ initialized at $1.0$ and exponentially decayed by a factor of 0.9 every 10 steps. The low-rank mapping matrix $\mathbf{L}$ is configured with a rank of $r=32$ and a learning rate of 0.5, optimized using Adam for 100 epochs on the constructed exemplar set.

\begin{table*}[t]
\centering
\caption{Performance comparison on ViT-L/14 across eight visual classification tasks. The best and second-best results among all compared merging methods are highlighted in \textbf{bold} and \underline{underlined}, respectively.}
\vspace{-5pt}
\label{tab:4}
\renewcommand{\arraystretch}{0.7}
\resizebox{0.85\textwidth}{!}{%
\begin{tabular}{c l c c c c c c c c c c}
\toprule
\multirow{2}{*}{Type} & \multirow{2}{*}{Method} & \multirow{2}{*}{Storage (MB)} & \multicolumn{9}{c}{Accuracy (\%)} \\
\cmidrule(lr){4-12}
 & & & SUN397 & Cars & RESISC45 & EuroSAT & SVHN & GTSRB & MNIST & DTD & AVG \\
\midrule
\multirow{3}{*}{-} & Pre-trained & -- & 66.87 & 77.86 & 71.33 & 62.26 & 58.43 & 50.55 & 76.31 & 55.43 & 64.88 \\
 & Traditional MTL & -- & 78.37 & 92.84 & 97.30 & 99.00 & 97.79 & 97.23 & 99.45 & 79.10 & 92.64 \\
 & Individual & -- & 84.86 & 92.34 & 97.37 & 99.74 & 98.12 & 99.24 & 99.69 & 84.15 & 94.44 \\
\midrule
\multirow{11}{*}{\rotatebox{90}{Static}} & Weight-Averaging \cite{WortsmanIGRLMNF22} & -- & 71.08 & 81.54 & 82.67 & 90.63 & 78.23 & 70.63 & 97.01 & 62.77 & 79.32 \\
 & Task-Arithmetic \cite{IlharcoRWSHF23} & -- & 73.90 & 82.14 & 86.67 & 92.74 & 87.91 & 86.78 & 98.93 & 65.64 & 84.34 \\
 & DARE \cite{Yu0Y0L24} & -- & 72.96 & 79.41 & 84.49 & 89.59 & 89.19 & 86.34 & 99.10 & 64.57 & 83.21 \\
 & Ties-Merging \cite{YadavTCRB23} & -- & 73.55 & 83.09 & 87.37 & 95.48 & 86.53 & 88.80 & 99.02 & 66.65 & 85.06 \\
 & Fisher Merging \cite{MatenaR22} & -- & 70.16 & 81.57 & 82.43 & 90.11 & 83.36 & 89.30 & 96.02 & 62.39 & 81.92 \\
 & DF-Merge \cite{LeeLWWCW25} & -- & 68.21 & 89.95 & 82.79 & 95.85 & 87.09 & 83.66 & 82.93 & 79.20 & 83.71 \\
 & RegMean \cite{Jin0P023} & -- & 73.74 & 87.17 & 90.29 & 97.93 & 92.60 & 91.26 & 99.21 & 71.76 & 88.00 \\
 & AdaMerging \cite{YangW00G0T24} & -- & 77.10 & 89.65 & 90.76 & 95.52 & 91.79 & 97.45 & 98.92 & 78.94 & 90.02 \\
 & AdaMerging++ \cite{YangW00G0T24} & -- & 77.48 & 90.48 & 90.71 & 96.59 & 92.76 & 97.36 & 99.23 & 78.98 & 90.45 \\
\midrule
\multirow{8}{*}{\rotatebox{90}{Dynamic}} & EMR-Merging \cite{huang2024emr} & 1633.46 & 83.17 & 90.71 & 96.76 & \underline{99.70} & 97.94 & 99.11 & 99.69 & 82.71 & 93.72 \\
 & Twin-Merging \cite{lu2024twin} & 1780.62 & 80.22 & 90.16 & 96.37 & 99.63 & \underline{98.06} & 98.95 & 99.70 & 82.29 & 93.17 \\
 & MoW-Merging \cite{YeHSCHO25} & 5159.51 & 80.99 & 91.32 & 96.25 & 99.59 & \underline{98.06} & 99.20 & \textbf{99.74} & 78.78 & 92.99 \\
 & Auto-Switch & 556.88 & 83.17 & \textbf{92.61} & 96.81 & \textbf{99.74} & \textbf{98.12} & 99.21 & \underline{99.73} & 83.51 & 94.11 \\
 & Auto-FlexSwitch+CKA ($\lambda=3$) & \underline{89.76} & 84.55 & 92.35 & \textbf{97.35} & \textbf{99.74} & 97.84 & \underline{99.23} & \textbf{99.74} & 84.26 & 94.38 \\
 & Auto-FlexSwitch+KL ($\lambda=0.3$) & \textbf{80.69} & \textbf{84.86} & \underline{92.58} & 97.19 & \underline{99.70} & 98.02 & \textbf{99.32} & 99.70 & \underline{84.47} & \textbf{94.48} \\
 & Auto-FlexSwitch+MSE ($\lambda=0.05$) & 98.41 & \underline{84.64} & 92.56 & \underline{97.33} & \textbf{99.74} & 98.04 & 99.22 & 99.69 & \textbf{84.57} & \underline{94.47} \\
\bottomrule
\end{tabular}%
}
\vspace{-10pt}
\end{table*}

In addition, for reference, we also compare against the pretrained model (Pre-trained), a model jointly trained on all datasets (Traditional MTL), and models individually fine-tuned on each task (Individual). For Traditional MTL, each model is trained for 20000 steps with a batch size of 64, using the Adam optimizer with a learning rate of 1e-4 for ViT-B/32 and ViT-L/14, and the SGD optimizer with a learning rate of 3e-2 and momentum of 0.9 for ConvNeXt. For the Individual setting, the fine-tuned weights for ViT-B/32 and ViT-L/14 are consistent with those in \cite{huang2024emr}, while the weights for ConvNeXt are optimized for 4,000 steps using SGD with a 3e-2 learning rate, a 64 batch size, and a momentum of 0.9.

\paragraph{Results}
The experimental results on the three backbones, ViT-B/32, ViT-L/14, and ConvNeXt, are presented in Tabs \ref{tab:3}, \ref{tab:4}, and \ref{tab:5}, respectively. Experimental results indicate that Auto-Switch consistently outperforms existing baselines in both performance and storage efficiency across ViT-B/32 and the larger-scale ViT-L/14. Auto-FlexSwitch further demonstrates superior overall competitiveness by achieving best average accuracy and even surpassing MTL and independent fine-tuning while significantly compressing additional storage. The storage advantage of Auto-FlexSwitch over Auto-Switch becomes even more pronounced on the larger ViT-L/14, yielding up to a 6.9$\times$ reduction in storage space. Notably, when the backbone is switched to the convolutional-based ConvNeXt, Auto-Switch struggles to maintain per-task performance. In contrast, leveraging the adaptive sparsification and quantization capabilities enabled by LGS and BAS, Auto-FlexSwitch sustains high performance on ConvNeXt while maintaining an exceptionally low storage footprint of 8.25 MB.

\begin{table*}[t]
\centering
\caption{Performance comparison on ConvNeXt across eight visual classification tasks. The best and second-best results among all compared merging methods are highlighted in \textbf{bold} and \underline{underlined}, respectively.}
\vspace{-5pt}
\label{tab:5}
\renewcommand{\arraystretch}{0.7}
\resizebox{0.85\textwidth}{!}{%
\begin{tabular}{c l c c c c c c c c c c}
\toprule
\multirow{2}{*}{Type} & \multirow{2}{*}{Method} & \multirow{2}{*}{Storage (MB)} & \multicolumn{9}{c}{Accuracy (\%)} \\
\cmidrule(lr){4-12}
 & & & SUN397 & Cars & RESISC45 & EuroSAT & SVHN & GTSRB & MNIST & DTD & AVG \\
\midrule
\multirow{3}{*}{-} & Pre-trained & -- & 54.12 & 10.43 & 59.89 & 77.56 & 22.55 & 37.43 & 63.27 & 62.82 & 48.51 \\
 & Traditional MTL & -- & 63.34 & 82.80 & 92.32 & 97.93 & 95.30 & 96.07 & 99.03 & 71.86 & 87.33 \\
 & Individual & -- & 66.95 & 88.26 & 95.71 & 99.00 & 97.06 & 97.77 & 99.35 & 72.93 & 89.63 \\
\midrule
\multirow{11}{*}{\rotatebox{90}{Static}} & Weight-Averaging \cite{WortsmanIGRLMNF22} & -- & 49.31 & 13.62 & 74.62 & 89.96 & 61.76 & 65.27 & 82.52 & 60.16 & 62.15 \\
 & Task-Arithmetic \cite{IlharcoRWSHF23} & -- & 42.75 & 13.31 & 71.41 & 90.63 & 73.16 & 72.78 & 92.84 & 53.40 & 63.79 \\
 & DARE \cite{Yu0Y0L24} & -- & 41.93 & 13.12 & 70.98 & 91.15 & 72.94 & 73.16 & 92.23 & 53.19 & 63.59 \\
 & Ties-Merging \cite{YadavTCRB23} & -- & 43.99 & 13.00 & 73.86 & 92.26 & 70.74 & 77.38 & 94.48 & 55.85 & 65.20 \\
 & Fisher Merging \cite{MatenaR22} & -- & 51.84 & 26.78 & 74.68 & 82.81 & 71.39 & 74.70 & 96.44 & 62.13 & 67.60 \\
 & DF-Merge \cite{LeeLWWCW25} & -- & 44.31 & 20.99 & 65.95 & 80.52 & 67.32 & 90.27 & 95.61 & 57.07 & 65.26 \\
 & RegMean \cite{Jin0P023} & -- & 51.10 & 26.68 & 78.81 & 90.52 & 76.14 & 78.67 & 94.05 & 61.76 & 69.72 \\
 & AdaMerging \cite{YangW00G0T24} & -- & 54.42 & 13.42 & 83.08 & 90.96 & 80.63 & 82.91 & 96.39 & 66.44 & 72.41 \\
 & AdaMerging++ \cite{YangW00G0T24} & -- & 57.51 & 20.07 & 84.29 & 89.07 & 81.00 & 83.59 & 96.26 & 67.45 & 71.03 \\
\midrule
\multirow{8}{*}{\rotatebox{90}{Dynamic}} & EMR-Merging \cite{huang2024emr} & 468.56 & 48.38 & 48.59 & 84.98 & 95.07 & 92.67 & 94.17 & 98.05 & 66.28 & 78.52 \\
 & Twin-Merging \cite{lu2024twin} & 548.38 & 64.70 & 86.81 & 90.95 & 97.56 & 96.78 & 97.28 & 99.26 & 72.38 & 88.22 \\
 & MoW-Merging \cite{YeHSCHO25} & 3014.75 & 65.58 & 87.07 & 94.11 & 87.11 & 96.25 & \textbf{97.64} & \textbf{99.30} & 72.29 & 87.42 \\
 & Auto-Switch & 167.76 & 53.93 & 70.55 & 85.38 & 73.37 & 92.44 & 93.48 & 98.89 & 63.67 & 78.96 \\
 & Auto-FlexSwitch+CKA ($\lambda=30$) & \underline{9.41} & 65.49 & 86.56 & 94.43 & \textbf{98.85} & \textbf{96.97} & 97.41 & 99.19 & 72.50 & 88.82 \\
 & Auto-FlexSwitch+KL ($\lambda=3$) & 13.65 & \textbf{65.86} & \textbf{87.61} & \underline{94.44} & 98.78 & \underline{96.95} & 97.45 & \underline{99.23} & \underline{72.52} & \underline{89.11} \\
 & Auto-FlexSwitch+MSE ($\lambda=0.5$) & \textbf{8.25} & \underline{65.68} & \underline{87.36} & \underline{94.79} & \underline{98.82} & 96.83 & \underline{97.52} & 99.22 & \textbf{72.82} & \textbf{89.13} \\
\bottomrule
\end{tabular}%
}
\vspace{-10pt}
\end{table*}

\begin{table}[t]
\centering
\caption{Hyperparameter settings for different methods on the DETR backbone. “--” indicates that the corresponding parameter is not applicable.}
\label{tab:6}
\renewcommand{\arraystretch}{0.6}
\vspace{-5pt}
\resizebox{0.4\textwidth}{!}{
\begin{tabular}{lcccc}
\toprule
\multirow{2}{*}{Methods} & \multicolumn{3}{c}{DETR} \\
\cmidrule(lr){2-4}
& Sparsity Rate & Sample Size & Scaling Coefficient \\
\midrule
Task-Arithmetic & -- & -- & 0.2 \\
DARE & 0.4 & -- & 0.2 \\
TIES-Merging & 0.1 & -- & 0.8 \\
Fisher Merging & -- & 7168 & -- \\
DF-Merge & -- & 12600 & -- \\
RegMean & -- & 11200 & -- \\
Twin-Merging & -- & 7000 & 0.2 \\
MoW-Merging & -- & 4568 & -- \\
Auto-Switch & 0.5 & 3500 & -- \\
Auto-FlexSwitch & -- & 3500 & -- \\
\bottomrule
\end{tabular}}
\vspace{-12pt}
\end{table}

\subsubsection{Merging on Object Detection Tasks}
\label{sec:img_obj}
To evaluate the effectiveness of our methods in complex perception tasks, we compare them with other baselines on various detection tasks.

\paragraph{Experimental Settings}
The evaluation employs object detection tasks from seven domains of the \textbf{R}obo\textbf{F}low \textbf{100} (RF100) dataset \cite{abs-2211-13523}, including Aerial, Videogames, Microscopic, Underwater, Documents, Electromagnetic, and Real World, to validate the generalization capability of our methods on heterogeneous detection tasks. These datasets span diverse fields and exhibit high semantic diversity. We employ the pretrained end-to-end object detector DETR-ResNet50 \cite{CarionMSUKZ20} as the backbone model and adopt mAP@50 as the evaluation metric for the corresponding tasks, while also measuring the additional parameter storage overhead (MB) introduced by each dynamic merging method. For all baselines involving randomness as well as our methods, the results are reported as the mean of three independent experiments.

\begin{table*}[t]
\centering
\caption{Performance comparison on DETR-ResNet50 across seven object detection tasks. The best and second-best results among all compared merging methods are highlighted in \textbf{bold} and \underline{underlined}, respectively.}
\vspace{-5pt}
\label{tab:7}
\renewcommand{\arraystretch}{0.7}
\resizebox{0.85\textwidth}{!}{%
\begin{tabular}{c l c c c c c c c c c}
\toprule
\multirow{2}{*}{Type} & \multirow{2}{*}{Method} & \multirow{2}{*}{Storage (MB)} & \multicolumn{8}{c}{mAP@50 (\%)} \\
\cmidrule(lr){4-11}
 & & & Aerial & Videogames & Microscopic & Underwater & Documents & Electromagnetic & Real World & AVG \\
\midrule
\multirow{3}{*}{-} & Pre-trained & -- & 0.13 & 1.70 & 0.29 & 0.21 & 0.09 & 1.20 & 1.26 & 0.70 \\
 & Traditional MTL & -- & 51.00 & 61.66 & 48.23 & 35.02 & 43.67 & 51.47 & 36.04 & 46.73 \\
 & Individual & -- & 53.35 & 65.10 & 54.83 & 35.26 & 46.74 & 53.37 & 41.27 & 49.99 \\
\midrule
\multirow{8}{*}{\rotatebox{90}{Static}} & Weight-Averaging \cite{WortsmanIGRLMNF22} & -- & 2.23 & 1.32 & 3.65 & 3.41 & 1.47 & 5.01 & 4.64 & 3.10 \\
 & Task-Arithmetic \cite{IlharcoRWSHF23} & -- & 3.86 & 1.04 & 3.18 & 3.77 & 1.81 & 4.74 & 4.16 & 3.22 \\
 & DARE \cite{Yu0Y0L24} & -- & 4.61 & 1.02 & 3.26 & 3.75 & 1.81 & 4.71 & 4.33 & 3.36 \\
 & Ties-Merging \cite{YadavTCRB23} & -- & 4.44 & 1.29 & 3.78 & 3.97 & 1.77 & 4.87 & 4.23 & 3.29 \\
 & Fisher Merging \cite{MatenaR22} & -- & 3.72 & 1.13 & 4.28 & 4.17 & 2.15 & 4.82 & 5.71 & 3.71 \\
 & DF-Merge \cite{LeeLWWCW25} & -- & 3.54 & 0.98 & 3.02 & 3.98 & 2.43 & 4.77 & 4.29 & 3.29 \\
 & RegMean \cite{Jin0P023} & -- & 2.47 & 1.17 & 3.91 & 3.49 & 1.54 & 5.47 & 6.27 & 3.47 \\
\midrule
\multirow{6}{*}{\rotatebox{90}{Dynamic}} & EMR-Merging \cite{huang2024emr} & 192.33 & 8.30 & 10.90 & 16.57 & 17.44 & 14.17 & 37.55 & 14.49 & 17.06 \\
 & Twin-Merging \cite{lu2024twin} & 313.98 & \underline{46.91} & 47.33 & 41.51 & \underline{26.20} & 36.19 & 41.99 & 26.24 & \underline{38.05} \\
 & MoW-Merging \cite{YeHSCHO25} & 1256.57 & 2.94 & 5.45 & 22.59 & 8.66 & 14.34 & 20.06 & \textbf{31.85} & 15.13 \\
 & Auto-Switch & \underline{65.48} & 42.96 & \underline{53.57} & \underline{42.17} & 23.94 & \underline{38.77} & \underline{43.63} & 19.74 & 37.83 \\
 & Auto-FlexSwitch+MSE ($\lambda=0.5$) & \textbf{46.97} & \textbf{48.51} & \textbf{60.89} & \textbf{44.50} & \textbf{26.75} & \textbf{40.97} & \textbf{46.40} & \underline{28.77} & \textbf{42.40} \\
\bottomrule
\end{tabular}%
}
\vspace{-10pt}
\end{table*}

\paragraph{Baselines}
We maintain the same baseline settings as described in Sec. \ref{sec:img_classify}. Since detection tasks involve not only object classification but also bounding box regression, we exclude AdaMerging and AdaMerging++ \cite{YangW00G0T24} from this comparison, as they are specifically designed for classification models and rely on class probability distributions. Meanwhile, for Auto-FlexSwitch, we only consider MSE as the performance preserving loss to ensure alignment of both the classification and regression branches. Furthermore, for both Auto-Switch and Auto-FlexSwitch, since DETR returns multiple object queries per sample, we first aggregate the KNN prediction probabilities for each sample via majority voting before performing the merging. For the training of Traditional MTL and Individual DETR models, we follow the original settings \cite{CarionMSUKZ20}, employing the AdamW optimizer with a batch size of 64 and learning rates of 1e-5 for the backbone and 1e-4 for the heads. Specifically, the Individual models are optimized for 15000 steps on each respective task, while the MTL model is trained for a total of 65000 steps to ensure convergence. Following the protocol in Sec. \ref{sec:img_classify}, we conduct a grid search for each merging method to determine its optimal hyperparameters, which are detailed in Tab. \ref{tab:6}.

\paragraph{Results}
Tab. \ref{tab:7} presents the comparative results of various model merging methods across seven cross-domain object detection tasks. First, static merging methods prove nearly ineffective in these scenarios, with the average mAP@50 failing to exceed 4\%. This indicates that parameter conflicts become highly sensitive when facing object detection tasks with significant domain shifts. In contrast, dynamic merging methods more effectively mitigate inter-task interference and preserve performance. Among them, Auto-FlexSwitch achieves the best average performance with the minimal additional storage overhead. Specifically, compared to the second-best method, Twin-Merging, Auto-FlexSwitch utilizes less than 15\% of the additional storage footprint (46.97 MB vs. 313.98 MB) yet delivers a 4.35\% higher average mAP (42.40\% vs. 38.05\%). Furthermore, it achieves top performance on the majority of tasks, strongly demonstrating the superiority of Auto-FlexSwitch in balancing detection accuracy and task vector compression efficiency.

\begin{table}[h]
\centering
\caption{Hyperparameter settings for different methods across RoBERTa and Mamba architectures. “--” indicates that the corresponding parameter is not applicable.}
\label{tab:8}
\renewcommand{\arraystretch}{0.6}
\vspace{-5pt}
\resizebox{0.45\textwidth}{!}{
\begin{tabular}{lccccccccc}
\toprule
\multirow{2}{*}{Methods} & \multicolumn{3}{c}{RoBERTa} & \multicolumn{3}{c}{Mamba} \\
\cmidrule(lr){2-4} \cmidrule(lr){5-7}
& $\alpha$ & $N$ & Scaling Coef & $\alpha$ & $N$ & Scaling Coef \\
\midrule
Task-Arithmetic & -- & -- & 0.2 & -- & -- & 0.3 \\
DARE & 0.3 & -- & 0.2 & 0.2 & -- & 0.4 \\
TIES-Merging & 0.8 & -- & 0.8 & 0.8 & -- & 0.8 \\
Fisher Merging & -- & 7168 & -- & -- & 7168 & -- \\
DF-Merge & -- & 12600 & -- & -- & 12600 & -- \\
RegMean & -- & 11200 & -- & -- & 11200 & -- \\
AdaMerging & -- & 14336 & -- & -- & 14336 & -- \\
AdaMerging++ & 0.7 & 14336 & -- & 0.5 & 14336 & -- \\
Twin-Merging & -- & 7000 & 0.2 & -- & 7000 & 0.2 \\
Auto-Switch & 0.5 & 2100 & -- & 0.5 & 2100 & -- \\
Auto-FlexSwitch & -- & 2100 & -- & -- & 2100 & -- \\
\bottomrule
\end{tabular}}
\vspace{-10pt}
\end{table}

\begin{table*}[t]
\centering
\caption{Performance comparison on RoBERTa-base across seven tasks from the GLUE benchmark. The best and second-best results among all compared merging methods are highlighted in \textbf{bold} and \underline{underlined}, respectively.}
\vspace{-5pt}
\label{tab:9}
\renewcommand{\arraystretch}{0.7}
\resizebox{0.8\textwidth}{!}{%
\begin{tabular}{c l c c c c c c c c c}
\toprule
\multirow{2}{*}{Type} & \multirow{2}{*}{Method} & \multirow{2}{*}{Storage (MB)} & MCC & \multicolumn{6}{c}{Accuracy} \\
\cmidrule(lr){5-10}
 & & & CoLA & SST-2 & MRPC & QQP & MNLI & QNLI & RTE & AVG \\
\midrule
\multirow{3}{*}{-} & Pre-trained & -- & 0.0000 & 0.4908 & 0.3162 & 0.3682 & 0.3182 & 0.5089 & 0.4729 & 0.3536  \\
 & Traditional MTL & -- & 0.5706 & 0.9404 & 0.8848 & 0.8819 & 0.8435 & 0.9136 & 0.7906 & 0.8322 \\
 & Individual & -- & 0.6018 & 0.9404 & 0.8922 & 0.9141 & 0.8717 & 0.9271 & 0.7906 & 0.8483 \\
\midrule
\multirow{10}{*}{\rotatebox{90}{Static}} & Weight-Averaging \cite{WortsmanIGRLMNF22} & -- & 0.1808 & 0.8188 & 0.7794 & 0.7960 & 0.4492 & 0.7106 & 0.6173 & 0.6217 \\
 & Task-Arithmetic \cite{IlharcoRWSHF23}& -- & 0.2909 & 0.8463 & 0.7843 & 0.8261 & 0.5741 & 0.7584 & 0.6462 & 0.6752 \\
 & DARE \cite{Yu0Y0L24}& -- & 0.2946 & 0.8498 & 0.7941 & 0.8254 & 0.5789 & 0.7569 & 0.6785 & 0.6826 \\
 & Ties-Merging \cite{YadavTCRB23}& -- & 0.3673 & 0.8727 & 0.7892 & 0.8296 & 0.6368 & 0.8069 & 0.7220 & 0.7178 \\
 & Fisher Merging \cite{MatenaR22}& -- & 0.1316 & 0.8383 & 0.7745 & 0.8026 & 0.4875 & 0.7366 & 0.6245 & 0.6279 \\
 & DF-Merge \cite{LeeLWWCW25}& -- & 0.4323 & 0.5929 & 0.7647 & 0.8019 & 0.6615 & 0.6658 & 0.5704 & 0.6414 \\
 & RegMean \cite{Jin0P023}& -- & 0.3388 & 0.9071 & 0.7647 & 0.8381 & 0.7034 & 0.8239 & 0.5812 & 0.7082 \\
 & AdaMerging \cite{YangW00G0T24}& -- & 0.1711 & 0.8784 & 0.7917 & 0.8365 & 0.6928 & 0.7238 & 0.6318 & 0.6752 \\
 & AdaMerging++ \cite{YangW00G0T24}& -- & 0.2981 & 0.8933 & 0.7843 & 0.8246 & 0.6465 & 0.7966 & 0.7292 & 0.7104 \\
\midrule
\multirow{7}{*}{\rotatebox{90}{Dynamic}} & EMR-merging \cite{huang2024emr}& 576.75 & 0.3934 & 0.9358 & 0.8701 & 0.8987 & 0.8578 & 0.8953 & 0.7473 & 0.7998 \\
 & Twin-merging \cite{lu2024twin}& 670.08 & 0.5651 & \underline{0.9392} & 0.8848 & 0.8820 & 0.8581 & \underline{0.9240} & 0.7437 & 0.8281 \\
 & Auto-Switch & 209.08 & 0.5339 & \textbf{0.9426} & \textbf{0.8921} & \textbf{0.9129} & 0.8547 & \textbf{0.9249} & 0.7509 & 0.8303 \\
 & Auto-FlexSwitch+CKA ($\lambda=5$) & 70.93 & \textbf{0.6121} & 0.9278 & \underline{0.8873} & 0.9100 & \textbf{0.8679} & 0.9215 & \underline{0.7545} & \textbf{0.8402} \\
 & Auto-FlexSwitch+KL ($\lambda=0.5$) & \underline{51.60} & 0.5806 & 0.9380 & 0.8856 & \underline{0.9105} & \underline{0.8674} & 0.9185 & \textbf{0.7581} & \underline{0.8370} \\
 & Auto-FlexSwitch+MSE ($\lambda=0.09$) & \textbf{35.70} & \underline{0.5830} & 0.9278 & 0.8824 & 0.9092 & 0.8644 & 0.9204 & 0.7445 & 0.8331 \\
\bottomrule
\end{tabular}%
}
\vspace{-10pt}
\end{table*}

\begin{table*}[t]
\centering
\caption{Performance comparison on Mamba-130M across seven tasks from the GLUE benchmark. The best and second-best results among all compared merging methods are highlighted in \textbf{bold} and \underline{underlined}, respectively.}
\vspace{-5pt}
\label{tab:10}
\renewcommand{\arraystretch}{0.7}
\resizebox{0.8\textwidth}{!}{%
\begin{tabular}{c l c c c c c c c c c}
\toprule
\multirow{2}{*}{Type} & \multirow{2}{*}{Method} & \multirow{2}{*}{Storage (MB)} & MCC & \multicolumn{6}{c}{Accuracy} \\
\cmidrule(lr){5-10}
 & & & CoLA & SST-2 & MRPC & QQP & MNLI & QNLI & RTE & AVG \\
\midrule
\multirow{3}{*}{-} & Pre-trained & -- & -0.0454 & 0.4920 & 0.6250 & 0.6220 & 0.3211 & 0.4939 & 0.5451 & 0.4362 \\
 & Traditional MTL & -- & 0.5168 & 0.9106 & 0.7721 & 0.8699 & 0.7833 & 0.8627 & 0.6679 & 0.7690 \\
 & Individual & -- & 0.5203 & 0.9197 & 0.7770 & 0.8864 & 0.8210 & 0.8794 & 0.6751 & 0.7827 \\
\midrule
\multirow{10}{*}{\rotatebox{90}{Static}} & Weight-Averaging \cite{WortsmanIGRLMNF22} & -- & 0.0352 & 0.7351 & 0.6299 & 0.6523 & 0.4077 & 0.5713 & 0.5993 & 0.5187 \\
 & Task-Arithmetic \cite{IlharcoRWSHF23}& -- & 0.1516 & 0.7477 & 0.6446 & 0.6165 & 0.5662 & 0.6661 & 0.6426 & 0.5765 \\
 & DARE \cite{Yu0Y0L24}& -- & 0.1384 & 0.8177 & 0.6005 & 0.5375 & 0.6376 & 0.6438 & 0.6318 & 0.5725 \\
 & Ties-Merging \cite{YadavTCRB23}& -- & 0.2015 & 0.7489 & 0.6740 & 0.6175 & 0.5534 & 0.6674 & 0.6534 & 0.5880 \\
 & Fisher Merging \cite{MatenaR22}& -- & 0.0980 & 0.7695 & 0.6446 & 0.6787 & 0.4340 & 0.5843 & 0.6029 & 0.5446 \\
 & DF-Merge \cite{LeeLWWCW25}& -- & 0.1423 & 0.8165 & 0.6936 & 0.4668 & 0.5954 & 0.5394 & 0.6137 & 0.5525 \\
 & RegMean \cite{Jin0P023}& -- & 0.1947 & 0.8876 & 0.6887 & 0.8066 & 0.5008 & 0.6877 & 0.6245 & 0.6272 \\
 & AdaMerging \cite{YangW00G0T24}& -- & 0.1566 & 0.6823 & 0.6446 & 0.5221 & 0.5983 & 0.6189 & 0.6170 & 0.5485 \\
 & AdaMerging++ \cite{YangW00G0T24}& -- & 0.2917 & 0.7546 & 0.6814 & 0.5426 & 0.5505 & 0.6442 & 0.6143 & 0.5828 \\
\midrule
\multirow{7}{*}{\rotatebox{90}{Dynamic}} & EMR-merging \cite{huang2024emr}& 600.37 & 0.4891 & 0.9117 & 0.7525 & 0.8383 & \textbf{0.8053} & 0.8495 & \textbf{0.6570} & 0.7576 \\
 & Twin-merging \cite{lu2024twin}& 795.29 & \underline{0.5084} & \underline{0.9174} & 0.7743 & 0.8759 & 0.7829 & 0.8706 & 0.6531 & 0.7689 \\
 & Auto-Switch & 216.46 & 0.4950 & \textbf{0.9186} & 0.7549 & 0.8786 & 0.7845 & 0.8424 & 0.6531 & 0.7610 \\
 & Auto-FlexSwitch+CKA ($\lambda=50$) & 61.60 & 0.5075 & 0.9163 & \underline{0.7768} & \textbf{0.8848} & \underline{0.8011} & \textbf{0.8742} & \underline{0.6534} & \textbf{0.7734} \\
 & Auto-FlexSwitch+KL ($\lambda=5$) & \underline{34.96} & 0.5068 & 0.9151 & \textbf{0.7892} & \underline{0.8812} & 0.7900 & \underline{0.8730} & 0.6498 & \underline{0.7722} \\
 & Auto-FlexSwitch+MSE ($\lambda=0.9$) & \textbf{23.90} & \textbf{0.5119} & 0.9128 & 0.7745 & 0.8783 & 0.7880 & 0.8699 & \underline{0.6534} & 0.7698 \\
\bottomrule
\end{tabular}%
}
\vspace{-10pt}
\end{table*}

\subsubsection{Merging on Natural Language Understanding Tasks}
\label{sec:language_und}
To validate the applicability of our method in the field of natural language processing, we compare the performance of different methods on natural language understanding tasks.

\paragraph{Experimental Settings}
The evaluation is conducted on seven classification tasks from the GLUE benchmark \cite{WangSMHLB19}, including CoLA \cite{WarstadtSB19}, SST-2 \cite{SocherPWCMNP13}, MRPC \cite{DolanB05}, QQP\footnote{\url{https://data.quora.com/First-Quora-Dataset-Release-Question-Pairs}}, MNLI \cite{WilliamsNB18}, QNLI \cite{RajpurkarZLL16}, and RTE \cite{GiampiccoloMDD07}. In terms of model architectures, we employ RoBERTa-base \cite{abs-1907-11692} with a Transformer architecture and Mamba-130M \cite{abs-2312-00752} based on the deep state space model. For evaluation metrics, we use accuracy for all tasks except CoLA, which is evaluated using the \textbf{M}atthews \textbf{C}orrelation \textbf{C}oefficient (MCC). For all methods involving randomness, results are averaged over three independent runs.

\paragraph{Baselines}
We adopt methods and settings largely consistent with those in Sec. \ref{sec:img_classify}. Since MoW-Merging is specifically tailored for vision tasks, we exclude it from this comparison. For the Individual scheme, we directly inherit the publicly available RoBERTa fine-tuning weights from \cite{huang2024emr}. As for the Mamba model, we employ the AdamW optimizer to fine-tune for 4000 steps with a batch size of 256 and a learning rate of $5e-5$. In the MTL setup, RoBERTa and Mamba are trained for 10 epochs using AdamW, with learning rates of $2e-5$ and $5e-5$, respectively. Similar to Sec. \ref{sec:img_classify}, we perform a grid search for each merging method to determine its optimal parameter configuration on each model, detailed settings are provided in Tab. \ref{tab:8}.

\paragraph{Results}
The results on RoBERTa and Mamba are presented in Tab. \ref{tab:9} and Tab. \ref{tab:10}, respectively. Auto-Switch and Auto-FlexSwitch consistently demonstrate competitive merging performance across both architectures. Notably, Auto-FlexSwitch achieves the best average performance with the minimal additional storage footprint on both models, even surpassing the performance of individually fine-tuned models on the CoLA task for RoBERTa and the MRPC task for Mamba. Benefiting from the adaptive task-vector compression enabled by the LGS and BAS mechanisms, Auto-FlexSwitch outperforms existing methods while requiring only 3\%--17\% of the additional storage compared to other dynamic merging approaches (including Auto-Switch). These results further confirm the effectiveness and generalizability of Auto-FlexSwitch in balancing performance and storage efficiency.

\begin{table}[t]
\centering
\caption{Component ablation analysis of Auto-FlexSwitch. Results formatted as 'xx/xx/xx' correspond to sparsity ratios $\alpha \in \{0.5, 0.7, 0.9\}$.}
\vspace{-5pt}
\label{tab:11}
\begin{adjustbox}{width=0.48\textwidth}
\begin{tabular}{lcccc}
\toprule
\multirow{2}{*}{Method} & \multicolumn{2}{c}{ViT-B/32} & \multicolumn{2}{c}{RoBERTa-base} \\
\cmidrule(lr){2-3} \cmidrule(lr){4-5}
 & AVG Performance & Storage & AVG Performance & Storage \\
\midrule
Auto-Switch & 90.81/90.78/85.46 & 216.96/152.08/72.00 & 0.8303/0.7982/0.5919 & 209.08/146.88/70.24 \\
Auto-Switch+LGS & 90.8 & 44.04 & 0.8267 & 55.22 \\
Auto-Switch+BAS & 91.05/91.02/89.41 & 220.29/152.93/72.2 & 0.8319/0.8171/0.7525 & 210.64/146.88/70.42 \\
Auto-Switch+LGS+BAS & 90.95 & 40.25 & 0.8309 & 53.34 \\
Auto-FlexSwitch & 91.03 & 40.31 & 0.8370 & 51.60 \\
\bottomrule
\end{tabular}
\end{adjustbox}
\vspace{-10pt}
\end{table}

\subsection{Ablation Study and Analysis}
In this section, we conduct ablation studies on the individual components, hyperparameters, and efficiency of the proposed method. Unless otherwise specified, all settings remain consistent with those in the main experiments (Sec. \ref{sec:main_results}). All results are reported as the average of three independent runs.
\paragraph{Component Analysis of Auto-FlexSwitch}
To investigate the contribution of each component within Auto-FlexSwitch, we evaluate the average performance and storage overhead across ViT-B/32 and RoBERTa-base under different configurations, including: 1) the vanilla Auto-Switch with sparsity ratios $\alpha \in \{0.5, 0.7, 0.9\}$; 2) Auto-Switch integrated with the LGS sparsification scheme; 3) Auto-Switch using the BAS quantization scheme (also at $\alpha \in \{0.5, 0.7, 0.9\}$); 4) Auto-Switch combined with both LGS and BAS; and 5) the full Auto-FlexSwitch. We uniformly set the exemplar size to 300 samples per task and employ KL divergence as the performance preservation loss. The results summarized in Tab. \ref{tab:11} indicate that the LGS mechanism significantly reduces storage overhead while maintaining performance levels comparable to the vanilla Auto-Switch at $\alpha=0.5$. Secondly, the introduction of BAS consistently enhances the performance of Auto-Switch across all sparsity ratios. Notably, the storage footprint of Auto-Switch+LGS+BAS is even lower than that of LGS alone. This phenomenon stems from the synergy between BAS and LGS: BAS relaxes the upper bound of the LGS sparsity ratio, allowing the system to converge to a lower storage cost (e.g., from 92.96\% to 94.95\% on ViT, and from 88.69\% to 91.73\% on RoBERTa). Finally, the KNN merging mechanism with a learnable low-rank metric achieves more accurate task retrieval to further boost the performance of Auto-FlexSwitch. These results fully validate the effectiveness of each component.

\begin{figure}[t]
    \centering
    \includegraphics[width=0.48\textwidth]{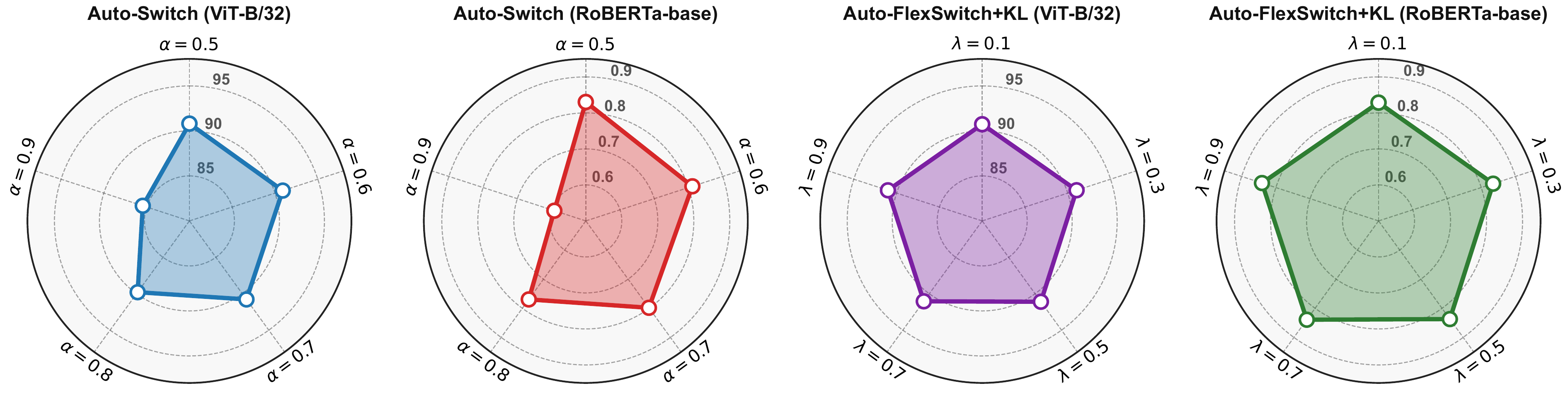}
    \vspace{-7pt}
    \caption{Comparison of model performance under different settings of the sparsity ratio $\alpha$ in Auto-Switch and the coefficient $\lambda$ in Auto-FlexSwitch.}
    \label{fig:ablation_SR}
    \vspace{-12pt}
\end{figure}

\paragraph{Sensitivity Analysis on Sparsity Ratios and Regularization Strengths}
We analyze the relationship of model performance with the sparsity ratio $\alpha$ in Auto-Switch and the performance preservation coefficient $\lambda$ in Auto-FlexSwitch to explore their respective impacts. Specifically, we record the average performance of Auto-Switch with $\alpha \in \{0.5, 0.6, 0.7, 0.8, 0.9\}$ and Auto-FlexSwitch with $\lambda \in \{0.1, 0.3, 0.5, 0.7, 0.9\}$ on ViT-B/32 and RoBERTa-base using 300 exemplars per task. As illustrated in Fig. \ref{fig:ablation_SR}, Auto-Switch is relatively sensitive to $\alpha$, and its performance exhibits a clear shrinking trend when $\alpha$ takes large values. Notably on RoBERTa-base, its performance drops sharply to 0.5919 when $\alpha=0.9$. In contrast, Auto-FlexSwitch demonstrates superior robustness across various $\lambda$ values. Since $\lambda$ directly regulates the weight of the performance preservation loss, increasing its value effectively strengthens the anchoring of core knowledge. Consequently, the performance of Auto-FlexSwitch on RoBERTa-base improves steadily with larger $\lambda$ and reaches 0.8405 at $\lambda=0.9$, while the performance on ViT-B/32 remains stable at a high level of approximately 91\%. In summary, Auto-FlexSwitch effectively suppresses performance degradation via the $\lambda$ to significantly enhance performance controllability during task vector compression.

\paragraph{Effect of Exemplar Size and Number of Nearest Neighbors}
\begin{figure}[t]
    \centering
    \includegraphics[width=0.48\textwidth]{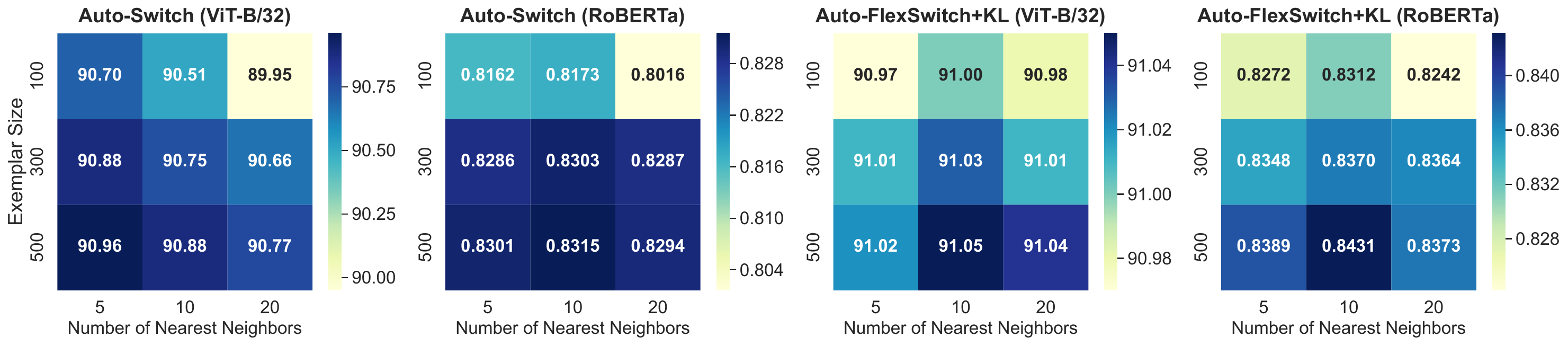}
    \vspace{-10pt}
    \caption{Performance of Auto-Switch and Auto-FlexSwitch under varying numbers of exemplars and neighbors.}
    \label{fig:ablation_Sample}
    \vspace{-12pt}
\end{figure}
To evaluate the impact of exemplar sample size and the number of neighbors on the performance of Auto-Switch and Auto-FlexSwitch, we test the average performance of both methods on ViT-B/32 and RoBERTa-base using exemplar sizes of $\{100, 300, 500\}$ and neighbor counts of $\{5, 10, 20\}$. As illustrated in Fig. \ref{fig:ablation_Sample}, Auto-FlexSwitch consistently outperforms Auto-Switch and demonstrates superior robustness across all parameter configurations. In terms of sample size, Auto-FlexSwitch shows low sensitivity to data scale; even under the constrained setting of only 100 samples, its performance on ViT-B/32 fluctuates minimally. Furthermore, both methods achieve stable performance when the number of neighbors is set to 10. In summary, Auto-FlexSwitch achieves effective dynamic merging with low data dependency.

\paragraph{Analysis of Low-rank Dimension and Center Quantity in Auto-FlexSwitch}
\begin{figure}[t]
    \centering
    \includegraphics[width=0.48\textwidth]{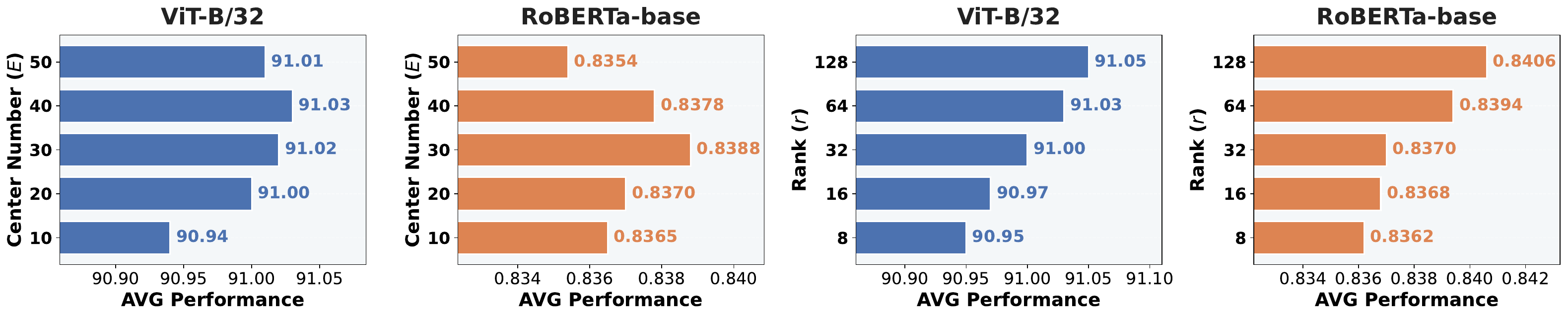}
    \vspace{-10pt}
    \caption{Ablation analysis of center number $E$ (left two panels) and low-rank dimension $r$ (right two panels) on the performance of Auto-FlexSwitch across ViT-B/32 and RoBERTa-base.}
    \label{fig:ablation_rank}
    \vspace{-12pt}
\end{figure}
To investigate the impact of the rank $r$ and the number of centroids $E$ in the learnable low-rank metric, we evaluate the average performance of Auto-FlexSwitch+KL on ViT-B/32 and RoBERTa-base across $r \in \{8, 16, 32, 64, 128\}$ and $E \in \{10, 20, 30, 40, 50\}$. Specifically, we fix $E=20$ when ablating $r$, and vice versa with $r=32$. As illustrated in Fig. \ref{fig:ablation_rank}, regarding the number of centroids, since insufficient centroids fail to cover full task characteristics while excessive ones may introduce noise, a moderate number ($E=20$–$40$) thus helps achieve stable and superior performance. As for the rank parameter $r$, the performance improves steadily as it increases, and notably, competitive results are already achievable even at a small rank of $r=8$.

\paragraph{Detailed Storage Analysis for SASS across Different Sparsity Ratios}
\begin{figure}[h]
    \centering
    \includegraphics[width=0.48\textwidth]{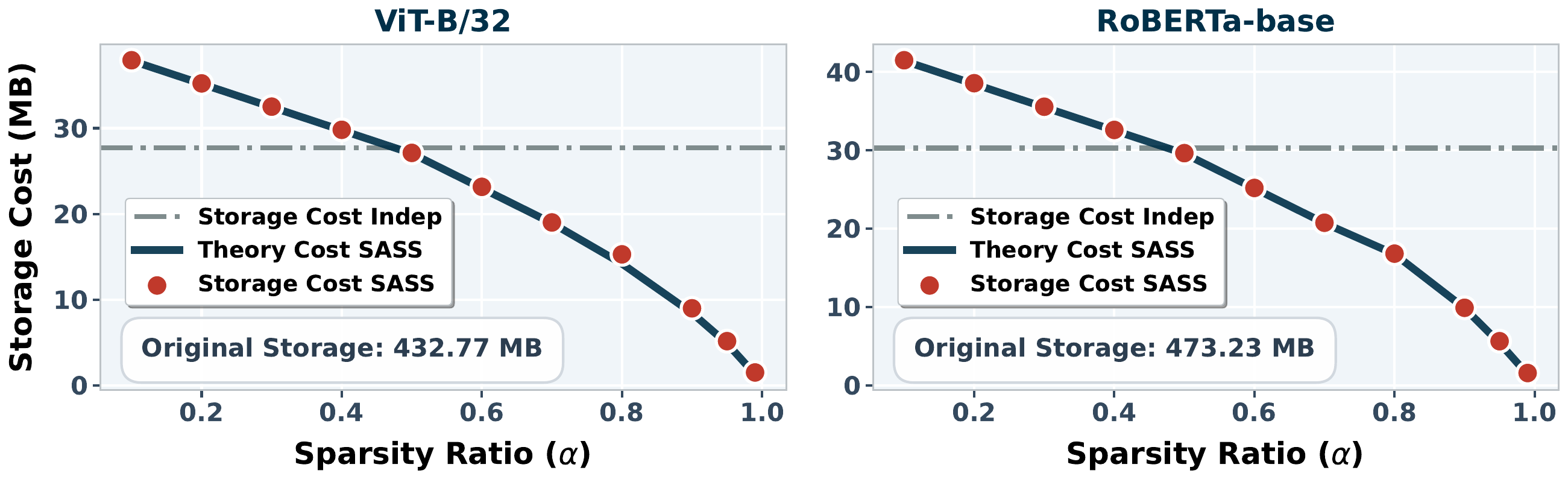}
    \vspace{-12pt}
    \caption{Storage overhead analysis of SASS under different sparsity ratios $\alpha$, where dots denote measured values, solid lines represent theoretical values, and the horizontal dashed line indicates the overhead of the Indep scheme.}
    \label{fig:ablation_sass}
    \vspace{-8pt}
\end{figure}
To quantitatively evaluate the efficiency of the SASS storage scheme, we leverage Auto-Switch to measure the actual storage overhead of individual task vectors for ViT-B/32 and RoBERTa-base across various sparsity ratios $\alpha \in \{0.1, 0.2, \dots, 0.99\}$. For comparison, we consider two baselines: full-precision storage (the original task vector) and the Indep storage scheme defined in Sec. \ref{sec:sass}, which independently stores the binarized masks, the quantized task vectors, and the scaling coefficients. Furthermore, we calculate the theoretical SASS storage overhead based on Eq. \eqref{eq:storage} to verify our analytical accuracy. As illustrated in Fig. \ref{fig:ablation_sass}, the empirical storage overhead of SASS aligns closely with the theoretical values from Eq. \eqref{eq:storage}, demonstrating strong alignment between theory and our physical implementation. When $\alpha > 0.5$, SASS begins to outperform the Indep scheme, with the efficiency gains increasing steadily as the sparsity ratio rises. Compared to the original full-precision weights, SASS achieves a substantial compression ratio. Specifically, for $\alpha > 0.9$, the compression ratio exceeds 40$\times$, and in extreme cases (e.g., $\alpha = 0.99$), it reaches over 200$\times$ (1.52MB on ViT and 1.59MB on RoBERTa), fully validating its storage efficiency for sparse and quantized task vectors. 

\begin{table}[t]
\centering
\caption{Comparison of training time (seconds) and average performance between Auto-FlexSwitch and MTL. 'A/B' denotes durations for the FlexSwitch stage and low-rank metric training, respectively.}
\renewcommand{\arraystretch}{0.7}
\vspace{-7pt}
\label{tab:training_time}
\begin{adjustbox}{width=0.48\textwidth}
\begin{tabular}{lcccc}
\toprule
\multirow{2}{*}{Method} & \multicolumn{2}{c}{ViT-B/32} & \multicolumn{2}{c}{RoBERTa-base} \\
\cmidrule(lr){2-3} \cmidrule(lr){4-5}
& Time & AVG Performance & Time & AVG Performance \\
\midrule
MTL & 17599.27 & 88.78 & 23520.59 & 83.22 \\
Auto-FlexSwitch+CKA & 1357.79 / 50.31 & 91.16 & 1001.72 / 85.49 & 84.02 \\
Auto-FlexSwitch+KL & 1355.34 / 52.09 & 91.00 & 967.32 / 84.64 & 83.70 \\
Auto-FlexSwitch+MSE & 1360.09 / 51.58 & 90.94 & 1001.05 / 84.26 & 83.31 \\
\bottomrule
\end{tabular}
\end{adjustbox}
\vspace{-10pt}
\end{table}

\begin{table}[t]
\centering
\caption{Inference time ratios of Auto-Switch and Auto-FlexSwitch relative to a single end-to-end inference, under varying exemplar counts and numbers of centroids.}
\renewcommand{\arraystretch}{0.7}
\vspace{-7pt}
\label{tab:inference_time}
\begin{adjustbox}{width=0.48\textwidth}
\begin{tabular}{lcccccc}
\toprule
\multirow{2}{*}{Model} & \multicolumn{3}{c}{Auto-Switch} & \multicolumn{3}{c}{Auto-FlexSwitch} \\
\cmidrule(lr){2-4} \cmidrule(lr){5-7}
& $N=100$ & $N=300$ & $N=500$ & $E=10$ & $E=30$ & $E=50$ \\
\midrule
ViT-B/32 & 2.27$\times$ & 2.55$\times$ & 2.82$\times$ & 1.88$\times$ & 1.90$\times$ & 1.91$\times$ \\
RoBERTa-base & 2.49$\times$ & 2.96$\times$ & 3.44$\times$ & 2.09$\times$ & 2.10$\times$ & 2.10$\times$ \\
\bottomrule
\end{tabular}
\end{adjustbox}
\vspace{-12pt}
\end{table}

\paragraph{Efficiency Analysis of Training and Inference}
To analyze the training and inference overheads of our proposed methods, we evaluate the training time of the sparsification-quantization process and the low-rank metric learning across different performance-preserving losses. For comparison, we also measure the training time and average performance of MTL. Furthermore, we record the average inference time on 1000 samples for Auto-Switch (with varying exemplar counts $N$) and Auto-FlexSwitch (with varying numbers of centroids $E$) on ViT-B/32 and RoBERTa-base, reporting their ratios relative to a single end-to-end inference. As shown in Tab. \ref{tab:training_time}, Auto-FlexSwitch incurs significantly lower training costs than MTL while achieving superior average performance; considering the massive reduction in storage overhead, its training efficiency is highly acceptable. Regarding inference efficiency, as shown in Tab. \ref{tab:inference_time}, the average inference cost of Auto-Switch is $2\times$ to $3\times$ that of a single end-to-end inference and escalates notably as the exemplar count $N$ increases. In contrast, benefiting from the enhanced retrieval efficiency of KNN in the learned low-rank space, Auto-FlexSwitch maintains a remarkably stable and lower inference cost even as the number of centroids $E$ grows. For instance, on RoBERTa-base, the inference time ratio of Auto-FlexSwitch remains steady at approximately 2.1$\times$, clearly outperforming the 2.49$\times$ of Auto-Switch at $N=100$. These results collectively demonstrate that Auto-FlexSwitch possesses favorable efficiency and scalability for both training and deployment.

\begin{table}[t]
  \centering
  \caption{Performance and storage comparison of FlexSwitch as a compression method for fine-tuned LLM incremental weights. Percentages denote the ratio of compressed storage to the original fine-tuned weight storage.}
  \renewcommand{\arraystretch}{0.7}
  \vspace{-5pt}
  \label{tab:ft_compress}
  \begin{adjustbox}{width=0.48\textwidth}
  \begin{tabular}{cccccccc}
    \toprule
    \textbf{Model} & \textbf{Storage (MB)} & \textbf{MATH} & \textbf{GSM8K} & \textbf{MBPP} & \textbf{TheoremQA} & \textbf{BBH} & \textbf{AVG} \\
    \midrule
    LLama-3.2-3B-Instruct &  12255.67 & 43.90 & 76.95 & 49.80 & 24.25 & 55.38 & 50.06 \\
    Bohdi-LLama-3.2-3B-Instruct & $\downarrow$  & 48.22 & 78.32 & 51.00 & 26.12 & 55.85 & 51.90 \\
    LLama-3.2-3B-Instruct + FlexSwitch & \textbf{5.21} (0.04\%) & 47.80 & 79.91 & 51.40 & 26.38 & 55.73 & 52.24 \\
    \cmidrule(r){1-8}
    gemma-2-9b-it & 35254.31 & 47.00 & 82.41 & 60.00 & 12.50 & 71.55 & 54.69 \\
    Bohdi-gemma-2-9b-it & $\downarrow$ & 53.76 & 84.31 & 59.20 & 25.75 & 73.69 & 59.34 \\
    gemma-2-9b-it + FlexSwitch & \textbf{731.57} (2.08\%) & 54.38 & 84.84 & 60.20 & 24.50 & 73.47 & 59.48 \\
    \bottomrule
  \end{tabular}
  \end{adjustbox}
  \vspace{-12pt}
\end{table}

\subsection{FlexSwitch as a Compression Method for Fine-Tuned LLM Weights}
Finally, we further explore the potential of FlexSwitch as a lightweight storage solution for fine-tuned incremental weights of LLMs. Specifically, we consider two LLMs with different scales and architectures, Llama-3.2-3B-Instruct \cite{abs-2407-21783} and gemma-2-9b-it \cite{abs-2408-00118}, along with their multi-domain fine-tuned counterparts, Bohdi-Llama-3.2-3B-Instruct \cite{gao2025bohdi} and Bohdi-gemma-2-9b-it \cite{gao2025bohdi}. We perform the FlexSwitch+KL training process on the LIMA dataset \cite{ZhouLX0SMMEYYZG23}, which contains 1030 samples, and evaluate the original model, the fine-tuned model, and the model loaded with FlexSwitch-compressed fine-tuned incremental weights on several representative reasoning and code generation benchmarks, including MATH \cite{HendrycksBKABTS21}, GSM8K \cite{abs-2110-14168}, MBPP \cite{abs-2108-07732}, TheoremQA \cite{ChenYKLWMXWX23}, and \textbf{B}IG-\textbf{B}ench \textbf{H}ard (BBH) \cite{SuzgunSSGTCCLCZ23}, all of which are disjoint from LIMA. For each configuration, we record the performance and the storage footprint of the incremental weights before and after compression. As shown in Tab. \ref{tab:ft_compress}, FlexSwitch effectively preserves fine-tuned performance on both LLMs with extremely low storage overhead. On Llama3.2-3B-Instruct, FlexSwitch requires only 5.21 MB (0.04\% of the original fine-tuned incremental weights) and achieves an average performance of 52.24\%, even slightly surpassing the original fine-tuned model, with improvements on GSM8K, MBPP and TheoremQA. On the larger Gemma-2-9b-it, FlexSwitch attains an average performance of 59.48\% with 731.57 MB of storage (merely 2.08\% of the original), again surpassing the fine-tuned model with substantially lower storage cost. These results validate the effectiveness of FlexSwitch as a lightweight storage scheme for fine-tuned incremental weights of LLMs. Notably, the FlexSwitch training here does not use any exemplar data from the same data source as the test data, nor the training data of the fine-tuned models \cite{gao2025bohdi}, indicating that its data dependency is not strictly confined to the same data source.

\section{Conclusions And Future Works}
Focusing on the storage efficiency bottlenecks in dynamic model merging, this work systematically explores efficient compression and adaptive composition mechanisms for task vectors. We first experimentally demonstrate that task vectors exhibit a notable impulse-like activation pattern and high robustness to low-bit representations, providing critical insights for designing lightweight task vectors. Based on this observation, we propose T-Switch, which achieves high-fidelity approximation at a high compression ratio via three components: a sparse mask, a sign vector and a scaling factor, and combines it with the adaptive merging scheme Auto-Switch to enable flexible combination of task knowledge. To further transcend the limitations of fixed rules, we designed FlexSwitch, a learnable framework for constructing lightweight task vectors. It utilizes the proposed LGS and BAS to jointly optimize the sparse structure and quantization bit-width of task vectors, while incorporating the SASS mechanism to automatically select the optimal storage encoding structure. Building upon this, by integrating a KNN mechanism with a learnable low-rank metric for inference-time merging, we develop Auto-FlexSwitch and validate its performance and efficiency across diverse scenarios and model architectures.

In future work, we will explore methods that directly encourage sparsity and quantizability during fine-tuning to further reduce training costs. We will also conduct a detailed investigation into the varying degrees of quantizability of task vectors corresponding to different components of mainstream model architectures, aiming to provide new insights for lightweight model knowledge storage. Additionally, we will investigate the feasibility of implementing online combinatorial decision-making for task vectors aligned with capability dimensions in more practical dynamic scenarios such as embodied settings by leveraging similar methodologies.

\bibliographystyle{IEEEtran}
\bibliography{IEEEabrv,references}

\begin{thebibliography}{10}
\providecommand{\url}[1]{#1}
\csname url@samestyle\endcsname
\providecommand{\newblock}{\relax}
\providecommand{\bibinfo}[2]{#2}
\providecommand{\BIBentrySTDinterwordspacing}{\spaceskip=0pt\relax}
\providecommand{\BIBentryALTinterwordstretchfactor}{4}
\providecommand{\BIBentryALTinterwordspacing}{\spaceskip=\fontdimen2\font plus
\BIBentryALTinterwordstretchfactor\fontdimen3\font minus \fontdimen4\font\relax}
\providecommand{\BIBforeignlanguage}[2]{{%
\expandafter\ifx\csname l@#1\endcsname\relax
\typeout{** WARNING: IEEEtran.bst: No hyphenation pattern has been}%
\typeout{** loaded for the language `#1'. Using the pattern for}%
\typeout{** the default language instead.}%
\else
\language=\csname l@#1\endcsname
\fi
#2}}
\providecommand{\BIBdecl}{\relax}
\BIBdecl

\bibitem{abs-1910-03771}
T.~Wolf, L.~Debut, V.~Sanh, J.~Chaumond, C.~Delangue, A.~Moi, P.~Cistac, T.~Rault, R.~Louf, M.~Funtowicz, and J.~Brew, ``Huggingface's transformers: State-of-the-art natural language processing,'' \emph{CoRR}, vol. abs/1910.03771, 2019.

\bibitem{abs-1906-07155}
K.~Chen, J.~Wang, J.~Pang, Y.~Cao, Y.~Xiong, X.~Li, S.~Sun, W.~Feng, Z.~Liu, J.~Xu, Z.~Zhang, D.~Cheng, C.~Zhu, T.~Cheng, Q.~Zhao, B.~Li, X.~Lu, R.~Zhu, Y.~Wu, J.~Dai, J.~Wang, J.~Shi, W.~Ouyang, C.~C. Loy, and D.~Lin, ``Mmdetection: Open mmlab detection toolbox and benchmark,'' \emph{CoRR}, vol. abs/1906.07155, 2019.

\bibitem{abs-2505-09388}
A.~Yang, A.~Li, B.~Yang, B.~Zhang, B.~Hui, B.~Zheng, B.~Yu, C.~Gao, C.~Huang, C.~Lv, C.~Zheng, D.~Liu, F.~Zhou, F.~Huang, F.~Hu, H.~Ge, H.~Wei, H.~Lin, J.~Tang, J.~Yang, J.~Tu, J.~Zhang, J.~Yang, J.~Yang, J.~Zhou, J.~Lin, K.~Dang, K.~Bao, K.~Yang, L.~Yu, L.~Deng, M.~Li, M.~Xue, M.~Li, P.~Zhang, P.~Wang, Q.~Zhu, R.~Men, R.~Gao, S.~Liu, S.~Luo, T.~Li, T.~Tang, W.~Yin, X.~Ren, X.~Wang, X.~Zhang, X.~Ren, Y.~Fan, Y.~Su, Y.~Zhang, Y.~Zhang, Y.~Wan, Y.~Liu, Z.~Wang, Z.~Cui, Z.~Zhang, Z.~Zhou, and Z.~Qiu, ``Qwen3 technical report,'' \emph{CoRR}, vol. abs/2505.09388, 2025.

\bibitem{abs-2503-19786}
G.~Team, ``Gemma 3 technical report,'' \emph{CoRR}, vol. abs/2503.19786, 2025.

\bibitem{abs-2412-08905}
M.~I. Abdin, J.~Aneja, H.~S. Behl, S.~Bubeck, R.~Eldan, S.~Gunasekar, M.~Harrison, R.~J. Hewett, M.~Javaheripi, P.~Kauffmann, J.~R. Lee, Y.~T. Lee, Y.~Li, W.~Liu, C.~C.~T. Mendes, A.~Nguyen, E.~Price, G.~de~Rosa, O.~Saarikivi, A.~Salim, S.~Shah, X.~Wang, R.~Ward, Y.~Wu, D.~Yu, C.~Zhang, and Y.~Zhang, ``Phi-4 technical report,'' \emph{CoRR}, vol. abs/2412.08905, 2024.

\bibitem{IlharcoRWSHF23}
G.~Ilharco, M.~T. Ribeiro, M.~Wortsman, L.~Schmidt, H.~Hajishirzi, and A.~Farhadi, ``Editing models with task arithmetic,'' in \emph{International Conference on Learning Representations}, 2023.

\bibitem{MatenaR22}
M.~Matena and C.~Raffel, ``Merging models with fisher-weighted averaging,'' in \emph{Annual Conference on Neural Information Processing Systems}, 2022.

\bibitem{Jin0P023}
X.~Jin, X.~Ren, D.~Preotiuc{-}Pietro, and P.~Cheng, ``Dataless knowledge fusion by merging weights of language models,'' in \emph{International Conference on Learning Representations}, 2023.

\bibitem{lu2024twin}
Z.~Lu, C.~Fan, W.~Wei, X.~Qu, D.~Chen, and Y.~Cheng, ``Twin-merging: Dynamic integration of modular expertise in model merging,'' \emph{Annual Conference on Neural Information Processing Systems}, vol.~37, pp. 78\,905--78\,935, 2024.

\bibitem{huang2024emr}
C.~Huang, P.~Ye, T.~Chen, T.~He, X.~Yue, and W.~Ouyang, ``Emr-merging: Tuning-free high-performance model merging,'' \emph{Annual Conference on Neural Information Processing Systems}, vol.~37, pp. 122\,741--122\,769, 2024.

\bibitem{QiLWGL0025}
B.~Qi, F.~Li, Z.~Wang, J.~Gao, D.~Li, P.~Ye, and B.~Zhou, ``Less is more: Efficient model merging with binary task switch,'' in \emph{Conference on Computer Vision and Pattern Recognition}, 2025, pp. 15\,265--15\,274.

\bibitem{DraxlerVSH18}
F.~Draxler, K.~Veschgini, M.~Salmhofer, and F.~A. Hamprecht, ``Essentially no barriers in neural network energy landscape,'' in \emph{International Conference on Machine Learning}, 2018, pp. 1308--1317.

\bibitem{GaripovIPVW18}
T.~Garipov, P.~Izmailov, D.~Podoprikhin, D.~P. Vetrov, and A.~G. Wilson, ``Loss surfaces, mode connectivity, and fast ensembling of dnns,'' in \emph{Annual Conference on Neural Information Processing Systems}, 2018, pp. 8803--8812.

\bibitem{FrankleD0C20}
J.~Frankle, G.~K. Dziugaite, D.~M. Roy, and M.~Carbin, ``Linear mode connectivity and the lottery ticket hypothesis,'' in \emph{International Conference on Machine Learning}, 2020, pp. 3259--3269.

\bibitem{WortsmanIGRLMNF22}
M.~Wortsman, G.~Ilharco, S.~Y. Gadre, R.~Roelofs, R.~G. Lopes, A.~S. Morcos, H.~Namkoong, A.~Farhadi, Y.~Carmon, S.~Kornblith, and L.~Schmidt, ``Model soups: averaging weights of multiple fine-tuned models improves accuracy without increasing inference time,'' in \emph{International Conference on Machine Learning}, 2022, pp. 23\,965--23\,998.

\bibitem{LeeLWWCW25}
S.~Lee, J.~Liu, Q.~Wang, J.~Wang, X.~Cai, and Y.~Wu, ``Dynamic fisher-weighted model merging via bayesian optimization,'' in \emph{Conference of the Nations of the Americas Chapter of the Association for Computational Linguistics}, 2025, pp. 4923--4935.

\bibitem{AinsworthHS23}
S.~K. Ainsworth, J.~Hayase, and S.~S. Srinivasa, ``Git re-basin: Merging models modulo permutation symmetries,'' in \emph{International Conference on Learning Representations}, 2023.

\bibitem{YangW00G0T24}
E.~Yang, Z.~Wang, L.~Shen, S.~Liu, G.~Guo, X.~Wang, and D.~Tao, ``Adamerging: Adaptive model merging for multi-task learning,'' in \emph{International Conference on Learning Representations}, 2024.

\bibitem{WangYSPK20}
H.~Wang, M.~Yurochkin, Y.~Sun, D.~S. Papailiopoulos, and Y.~Khazaeni, ``Federated learning with matched averaging,'' in \emph{International Conference on Learning Representations}, 2020.

\bibitem{StoicaBBRHH24}
G.~Stoica, D.~Bolya, J.~Bjorner, P.~Ramesh, T.~Hearn, and J.~Hoffman, ``Zipit! merging models from different tasks without training,'' in \emph{International Conference on Learning Representations}, 2024.

\bibitem{Yu0Y0L24}
L.~Yu, B.~Yu, H.~Yu, F.~Huang, and Y.~Li, ``Language models are super mario: Absorbing abilities from homologous models as a free lunch,'' in \emph{International Conference on Machine Learning}, 2024.

\bibitem{YadavTCRB23}
P.~Yadav, D.~Tam, L.~Choshen, C.~A. Raffel, and M.~Bansal, ``Ties-merging: Resolving interference when merging models,'' in \emph{Annual Conference on Neural Information Processing Systems}, 2023.

\bibitem{YeHSCHO25}
P.~Ye, C.~Huang, M.~Shen, T.~Chen, Y.~Huang, and W.~Ouyang, ``Dynamic model merging with mixture of weights,'' \emph{{IEEE} Trans. Circuits Syst. Video Technol.}, vol.~35, no.~8, pp. 7925--7939, 2025.

\bibitem{GuoRK20}
D.~Guo, A.~M. Rush, and Y.~Kim, ``Parameter-efficient transfer learning with diff pruning,'' in \emph{Annual Meeting of the Association for Computational Linguistics}, 2021, pp. 4884--4896.

\bibitem{AnsellPKV22}
A.~Ansell, E.~M. Ponti, A.~Korhonen, and I.~Vulic, ``Composable sparse fine-tuning for cross-lingual transfer,'' in \emph{Annual Meeting of the Association for Computational Linguistics}, 2022, pp. 1778--1796.

\bibitem{HuZDWWLS22}
S.~Hu, Z.~Zhang, N.~Ding, Y.~Wang, Y.~Wang, Z.~Liu, and M.~Sun, ``Sparse structure search for delta tuning,'' in \emph{Annual Conference on Neural Information Processing Systems}, 2022.

\bibitem{SungNR21}
Y.~Sung, V.~Nair, and C.~Raffel, ``Training neural networks with fixed sparse masks,'' in \emph{Annual Conference on Neural Information Processing Systems}, 2021, pp. 24\,193--24\,205.

\bibitem{BhardwajPPGK0BW24}
K.~Bhardwaj, N.~P. Pandey, S.~Priyadarshi, V.~Ganapathy, S.~Kadambi, R.~Esteves, S.~Borse, P.~N. Whatmough, R.~Garrepalli, M.~van Baalen, H.~Teague, and M.~Nagel, ``Sparse high rank adapters,'' in \emph{Annual Conference on Neural Information Processing Systems}, 2024.

\bibitem{HeLJM25}
H.~He, J.~B. Li, X.~Jiang, and H.~Miller, ``{SMT:} fine-tuning large language models with sparse matrices,'' in \emph{International Conference on Learning Representations}, 2025.

\bibitem{NingW0Z0CLZ24}
W.~Ning, J.~Wang, Q.~Qi, M.~Zhu, H.~Sun, D.~Cheng, J.~Liao, and C.~Zhang, ``Fm-delta: Lossless compression for storing massive fine-tuned foundation models,'' in \emph{Annual Conference on Neural Information Processing Systems}, 2024.

\bibitem{LiuXLL0DC24}
J.~Liu, G.~Xiao, K.~Li, J.~D. Lee, S.~Han, T.~Dao, and T.~Cai, ``Bitdelta: Your fine-tune may only be worth one bit,'' in \emph{Annual Conference on Neural Information Processing Systems}, 2024.

\bibitem{ZhuG18}
M.~Zhu and S.~Gupta, ``To prune, or not to prune: Exploring the efficacy of pruning for model compression,'' in \emph{International Conference on Learning Representations}, 2018.

\bibitem{Sun0BK24}
M.~Sun, Z.~Liu, A.~Bair, and J.~Z. Kolter, ``A simple and effective pruning approach for large language models,'' in \emph{International Conference on Learning Representations}, 2024.

\bibitem{RadfordKHRGASAM21}
A.~Radford, J.~W. Kim, C.~Hallacy, A.~Ramesh, G.~Goh, S.~Agarwal, G.~Sastry, A.~Askell, P.~Mishkin, J.~Clark, G.~Krueger, and I.~Sutskever, ``Learning transferable visual models from natural language supervision,'' in \emph{International Conference on Machine Learning}, vol. 139, 2021, pp. 8748--8763.

\bibitem{XiaoHEOT10}
J.~Xiao, J.~Hays, K.~A. Ehinger, A.~Oliva, and A.~Torralba, ``{SUN} database: Large-scale scene recognition from abbey to zoo,'' in \emph{Conference on Computer Vision and Pattern Recognition}, 2010, pp. 3485--3492.

\bibitem{Krause0DF13}
J.~Krause, M.~Stark, J.~Deng, and L.~Fei{-}Fei, ``3d object representations for fine-grained categorization,'' in \emph{International Conference on Computer Vision Workshops}, 2013, pp. 554--561.

\bibitem{ChengHL17}
G.~Cheng, J.~Han, and X.~Lu, ``Remote sensing image scene classification: Benchmark and state of the art,'' \emph{Proc. {IEEE}}, vol. 105, no.~10, pp. 1865--1883, 2017.

\bibitem{HelberBDB19}
P.~Helber, B.~Bischke, A.~Dengel, and D.~Borth, ``Eurosat: {A} novel dataset and deep learning benchmark for land use and land cover classification,'' \emph{{IEEE} J. Sel. Top. Appl. Earth Obs. Remote. Sens.}, vol.~12, no.~7, pp. 2217--2226, 2019.

\bibitem{netzer2011reading}
Y.~Netzer, T.~Wang, A.~Coates, A.~Bissacco, B.~Wu, A.~Y. Ng \emph{et~al.}, ``Reading digits in natural images with unsupervised feature learning,'' in \emph{Annual Conference on Neural Information Processing Systems Workshop}, vol. 2011, no.~5, 2011, p.~7.

\bibitem{StallkampSSI11}
J.~Stallkamp, M.~Schlipsing, J.~Salmen, and C.~Igel, ``The german traffic sign recognition benchmark: {A} multi-class classification competition,'' in \emph{International Joint Conference on Neural Networks}, 2011, pp. 1453--1460.

\bibitem{Deng12}
L.~Deng, ``The {MNIST} database of handwritten digit images for machine learning research [best of the web],'' \emph{{IEEE} Signal Process. Mag.}, vol.~29, no.~6, pp. 141--142, 2012.

\bibitem{CimpoiMKMV14}
M.~Cimpoi, S.~Maji, I.~Kokkinos, S.~Mohamed, and A.~Vedaldi, ``Describing textures in the wild,'' in \emph{Conference on Computer Vision and Pattern Recognition}, 2014, pp. 3606--3613.

\bibitem{ZhuHMD17}
C.~Zhu, S.~Han, H.~Mao, and W.~J. Dally, ``Trained ternary quantization,'' in \emph{International Conference on Learning Representations}, 2017.

\bibitem{DaiDHSCW22}
D.~Dai, L.~Dong, Y.~Hao, Z.~Sui, B.~Chang, and F.~Wei, ``Knowledge neurons in pretrained transformers,'' in \emph{Annual Meeting of the Association for Computational Linguistics}, 2022, pp. 8493--8502.

\bibitem{xu2020knowledge}
G.~Xu, Z.~Liu, X.~Li, and C.~C. Loy, ``Knowledge distillation meets self-supervision,'' in \emph{European Conference on Computer Vision}, 2020, pp. 588--604.

\bibitem{guo2020online}
Q.~Guo, X.~Wang, Y.~Wu, Z.~Yu, D.~Liang, X.~Hu, and P.~Luo, ``Online knowledge distillation via collaborative learning,'' in \emph{Conference on Computer Vision and Pattern Recognition}, 2020, pp. 11\,020--11\,029.

\bibitem{yang2023online}
C.~Yang, Z.~An, H.~Zhou, F.~Zhuang, Y.~Xu, and Q.~Zhang, ``Online knowledge distillation via mutual contrastive learning for visual recognition,'' \emph{IEEE Transactions on Pattern Analysis and Machine Intelligence}, vol.~45, no.~8, pp. 10\,212--10\,227, 2023.

\bibitem{kullback1951information}
S.~Kullback and R.~A. Leibler, ``On information and sufficiency,'' \emph{The Annals of Mathematical Statistics}, vol.~22, no.~1, pp. 79--86, 1951.

\bibitem{cho2019efficacy}
J.~H. Cho and B.~Hariharan, ``On the efficacy of knowledge distillation,'' in \emph{International Conference on Computer Vision}, 2019, pp. 4794--4802.

\bibitem{zhao2022decoupled}
B.~Zhao, Q.~Cui, R.~Song, Y.~Qiu, and J.~Liang, ``Decoupled knowledge distillation,'' in \emph{Conference on Computer Vision and Pattern Recognition}, 2022, pp. 11\,953--11\,962.

\bibitem{0003MWFDX22}
Z.~Liu, H.~Mao, C.~Wu, C.~Feichtenhofer, T.~Darrell, and S.~Xie, ``A convnet for the 2020s,'' in \emph{Conference on Computer Vision and Pattern Recognition}, 2022, pp. 11\,966--11\,976.

\bibitem{abs-2211-13523}
F.~Ciaglia, F.~S. Zuppichini, P.~Guerrie, M.~McQuade, and J.~Solawetz, ``Roboflow 100: {A} rich, multi-domain object detection benchmark,'' \emph{CoRR}, vol. abs/2211.13523, 2022.

\bibitem{CarionMSUKZ20}
N.~Carion, F.~Massa, G.~Synnaeve, N.~Usunier, A.~Kirillov, and S.~Zagoruyko, ``End-to-end object detection with transformers,'' in \emph{European Conference on Computer Vision}, 2020, pp. 213--229.

\bibitem{WangSMHLB19}
A.~Wang, A.~Singh, J.~Michael, F.~Hill, O.~Levy, and S.~R. Bowman, ``{GLUE:} {A} multi-task benchmark and analysis platform for natural language understanding,'' in \emph{International Conference on Learning Representations}, 2019.

\bibitem{WarstadtSB19}
A.~Warstadt, A.~Singh, and S.~R. Bowman, ``Neural network acceptability judgments,'' \emph{Trans. Assoc. Comput. Linguistics}, vol.~7, pp. 625--641, 2019.

\bibitem{SocherPWCMNP13}
R.~Socher, A.~Perelygin, J.~Wu, J.~Chuang, C.~D. Manning, A.~Y. Ng, and C.~Potts, ``Recursive deep models for semantic compositionality over a sentiment treebank,'' in \emph{Conference on Empirical Methods in Natural Language Processing}, 2013, pp. 1631--1642.

\bibitem{DolanB05}
W.~B. Dolan and C.~Brockett, ``Automatically constructing a corpus of sentential paraphrases,'' in \emph{International Workshop on Paraphrasing}.\hskip 1em plus 0.5em minus 0.4em\relax Asian Federation of Natural Language Processing, 2005.

\bibitem{WilliamsNB18}
A.~Williams, N.~Nangia, and S.~R. Bowman, ``A broad-coverage challenge corpus for sentence understanding through inference,'' in \emph{Conference of the North American Chapter of the Association for Computational Linguistics}, 2018, pp. 1112--1122.

\bibitem{RajpurkarZLL16}
P.~Rajpurkar, J.~Zhang, K.~Lopyrev, and P.~Liang, ``Squad: 100, 000+ questions for machine comprehension of text,'' in \emph{Conference on Empirical Methods in Natural Language Processing}, 2016, pp. 2383--2392.

\bibitem{GiampiccoloMDD07}
D.~Giampiccolo, B.~Magnini, I.~Dagan, and B.~Dolan, ``The third {PASCAL} recognizing textual entailment challenge,'' in \emph{Proceedings of the ACL-PASCAL@ACL 2007 Workshop on Textual Entailment and Paraphrasing}, 2007, pp. 1--9.

\bibitem{abs-1907-11692}
Y.~Liu, M.~Ott, N.~Goyal, J.~Du, M.~Joshi, D.~Chen, O.~Levy, M.~Lewis, L.~Zettlemoyer, and V.~Stoyanov, ``Roberta: {A} robustly optimized {BERT} pretraining approach,'' \emph{CoRR}, vol. abs/1907.11692, 2019.

\bibitem{abs-2312-00752}
A.~Gu and T.~Dao, ``Mamba: Linear-time sequence modeling with selective state spaces,'' \emph{CoRR}, vol. abs/2312.00752, 2023.

\bibitem{abs-2407-21783}
L.~Team, ``The llama 3 herd of models,'' \emph{CoRR}, vol. abs/2407.21783, 2024.

\bibitem{abs-2408-00118}
G.~Team, ``Gemma 2: Improving open language models at a practical size,'' \emph{CoRR}, vol. abs/2408.00118, 2024.

\bibitem{gao2025bohdi}
J.~Gao, Z.~Guo, D.~Zhang, D.~Li, R.~Liu, P.~Li, K.~Tian, and B.~Qi, ``Bohdi: Heterogeneous llm fusion with automatic data exploration,'' in \emph{Annual Conference on Neural Information Processing Systems}, 2025.

\bibitem{ZhouLX0SMMEYYZG23}
C.~Zhou, P.~Liu, P.~Xu, S.~Iyer, J.~Sun, Y.~Mao, X.~Ma, A.~Efrat, P.~Yu, L.~Yu, S.~Zhang, G.~Ghosh, M.~Lewis, L.~Zettlemoyer, and O.~Levy, ``{LIMA:} less is more for alignment,'' in \emph{Annual Conference on Neural Information Processing Systems}, 2023.

\bibitem{HendrycksBKABTS21}
D.~Hendrycks, C.~Burns, S.~Kadavath, A.~Arora, S.~Basart, E.~Tang, D.~Song, and J.~Steinhardt, ``Measuring mathematical problem solving with the {MATH} dataset,'' in \emph{Proceedings of the Neural Information Processing Systems Track on Datasets and Benchmarks}, 2021.

\bibitem{abs-2110-14168}
K.~Cobbe, V.~Kosaraju, M.~Bavarian, M.~Chen, H.~Jun, L.~Kaiser, M.~Plappert, J.~Tworek, J.~Hilton, R.~Nakano, C.~Hesse, and J.~Schulman, ``Training verifiers to solve math word problems,'' \emph{CoRR}, vol. abs/2110.14168, 2021.

\bibitem{abs-2108-07732}
J.~Austin, A.~Odena, M.~I. Nye, M.~Bosma, H.~Michalewski, D.~Dohan, E.~Jiang, C.~J. Cai, M.~Terry, Q.~V. Le, and C.~Sutton, ``Program synthesis with large language models,'' \emph{CoRR}, vol. abs/2108.07732, 2021.

\bibitem{ChenYKLWMXWX23}
W.~Chen, M.~Yin, M.~Ku, P.~Lu, Y.~Wan, X.~Ma, J.~Xu, X.~Wang, and T.~Xia, ``Theoremqa: {A} theorem-driven question answering dataset,'' in \emph{Conference on Empirical Methods in Natural Language Processing}, 2023.

\bibitem{SuzgunSSGTCCLCZ23}
M.~Suzgun, N.~Scales, N.~Sch{\"{a}}rli, S.~Gehrmann, Y.~Tay, H.~W. Chung, A.~Chowdhery, Q.~V. Le, E.~H. Chi, D.~Zhou, and J.~Wei, ``Challenging big-bench tasks and whether chain-of-thought can solve them,'' in \emph{Findings of the Association for Computational Linguistics}, 2023.

\end{thebibliography}

\vfill

\end{document}